\pdfoutput=1
\documentclass[11pt]{article}

 \usepackage[final]{acl}

\usepackage{times}
\usepackage{latexsym}

\usepackage[T1]{fontenc}
\usepackage[utf8]{inputenc}

\usepackage{microtype}

\usepackage{inconsolata}

\usepackage{graphicx}
\usepackage{amsmath}
\usepackage{amssymb}
\usepackage{multirow}
\usepackage{booktabs} % For better table rules
\usepackage{caption}  % Optional, for caption customization
\usepackage{booktabs}   % For improved table rules
\usepackage{caption}    % For caption customization
\usepackage{makecell}   % For multi-line cells
\usepackage{float}
\usepackage{lipsum}
\usepackage{stfloats} % Add this to your preamble
\usepackage{xcolor}     % For color support
\usepackage{xcolor,colortbl}
\definecolor{gold1}{RGB}{255,233,150}   % 1st
\definecolor{gold2}{RGB}{255,244,200}   % 2nd
\definecolor{gold3}{RGB}{255,249,220}   % 3rd

\newcommand{\first}[1]{\cellcolor{gold1}\textbf{#1}}
\newcommand{\second}[1]{\cellcolor{gold2}#1}
\newcommand{\third}[1]{\cellcolor{gold3}#1}

\definecolor{citecolor}{HTML}{0071bc}
\definecolor{pinegreen}{rgb}{0.0, 0.47, 0.44}
\definecolor{cornellred}{rgb}{0.7, 0.11, 0.11}
\definecolor{cadmiumgreen}{rgb}{0.0, 0.42, 0.24}
\definecolor{royalblue}{rgb}{0.0, 0.14, 0.4}
\definecolor{spirodiscoball}{rgb}{0.06, 0.75, 0.99}
\definecolor{mylightblue}{rgb}{0.85, 0.90, 0.94}
\definecolor{kaistblue}{RGB}{20,135,200}
\definecolor{auburn}{RGB}{166,38,57}

\title{Mind the Blind Spots: \\ A Focus-Level Evaluation Framework for LLM Reviews}

\author{
    Hyungyu Shin$^{\dagger}$, Jingyu Tang$^{\ddagger}$, Yoonjoo Lee$^{\dagger}$, Nayoung Kim$^{\dagger}$, Hyunseung Lim$^{\dagger}$, \\
    \textbf{Ji Yong Cho$^{\S}$}, \textbf{Hwajung Hong$^{\dagger}$}, \textbf{Moontae Lee$^{\S,\|}$}, \textbf{Juho Kim$^{\dagger}$} \\\\
    $^{\dagger}$KAIST \quad $^{\ddagger}$Huazhong University of Science and Technology \\
    $^{\S}$LG AI Research \quad $^{\|}$University of Illinois Chicago \\\\
}

\begin{document}
\maketitle
\begin{abstract}
Peer review underpins scientific progress, but it is increasingly strained by reviewer shortages and growing workloads. Large Language Models (LLMs) can automatically draft reviews now, but determining whether LLM-generated reviews are trustworthy requires systematic evaluation. Researchers have evaluated LLM reviews at either surface-level (e.g., BLEU and ROUGE) or content-level (e.g., specificity and factual accuracy). Yet it remains uncertain whether LLM-generated reviews attend to the same critical facets that human experts weigh---the strengths and weaknesses that ultimately drive an accept-or-reject decision. We introduce a focus-level evaluation framework that operationalizes the focus as a normalized distribution of attention across predefined facets in paper reviews. Based on the framework, we developed an automatic focus-level evaluation pipeline based on two sets of facets: target (e.g., problem, method, and experiment) and aspect (e.g., validity, clarity, and novelty), leveraging 676 paper reviews\footnote{\url{https://figshare.com/s/d5adf26c802527dd0f62}} from OpenReview that consists of 3,657 strengths and weaknesses identified from human experts. The comparison of focus distributions between LLMs and human experts showed that the off-the-shelf LLMs consistently have a more biased focus towards examining technical validity while significantly overlooking novelty assessment when criticizing papers.

% the developed an automatic evaluation pipeline to assess the LLMs' paper review capability by comparing them with expert-generated reviews. By constructing a dataset\footnote{\url{https://figshare.com/s/d5adf26c802527dd0f62}} consisting of 676 OpenReview papers, we examined the agreement between LLMs and experts in their strength and weakness identifications. The results showed that LLMs lack balanced perspectives, significantly overlook novelty assessment when criticizing, and produce poor acceptance decisions. Our automated pipeline enables a scalable evaluation of LLMs' paper review capability over time.

\end{abstract}

\section{Introduction}

\begin{figure}[t]
\centering
\includegraphics[width=\linewidth]{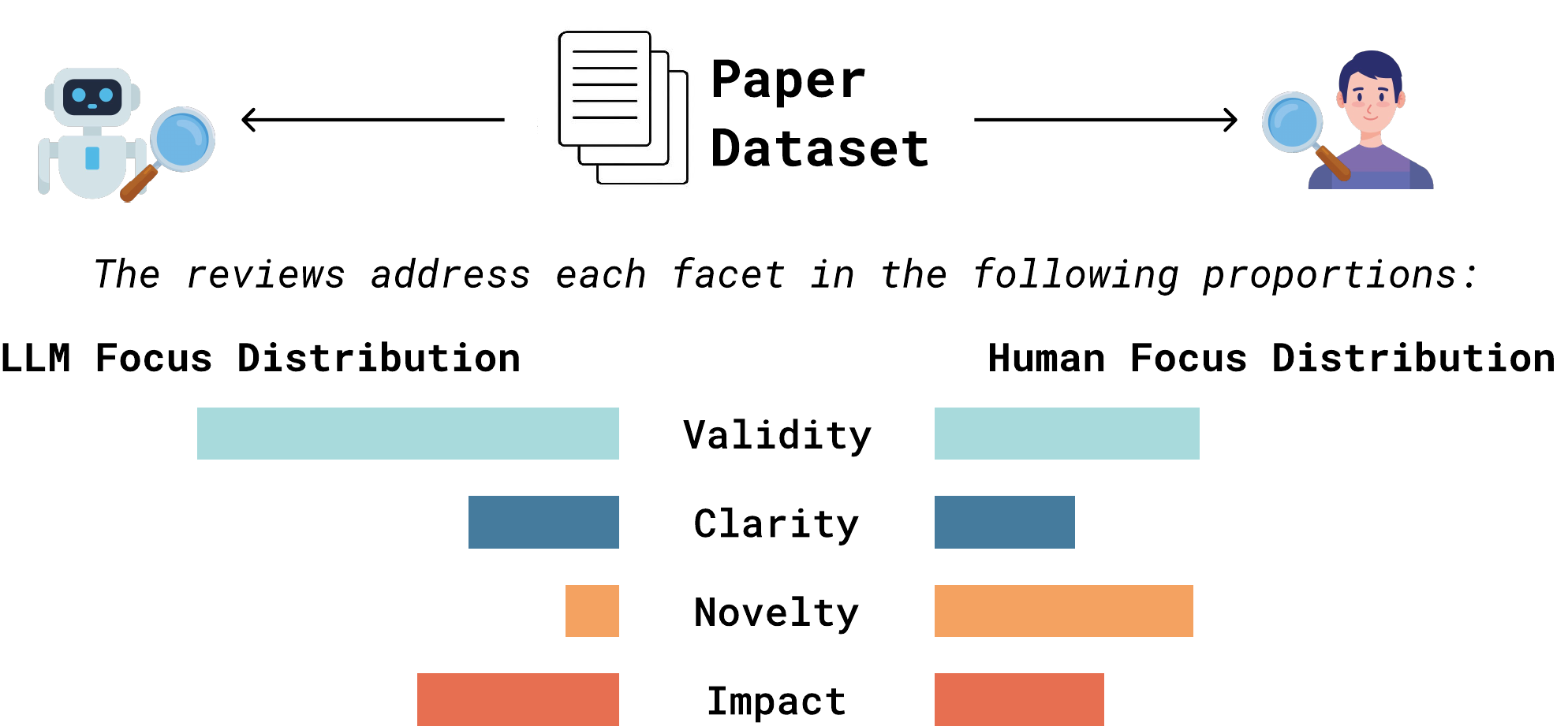}
\caption{We introduce a focus-level evaluation framework for assessing LLM reviews, which computes focus distributions and compares them against human reviews based on predefined facets. The focus-level evaluation offers actionable insights into how to improve LLMs' paper review capability and how to most effectively leverage LLM reviews in the peer review process.}
%We consider paper review task that generates a summary of paper, strengths, weaknesses, and the final judgement. Our goal is to examine the level of agreement between LLMs and human experts in reviewing papers, based on their feedback targets and aspects.
\label{fig:task}
\end{figure}

Reviewing academic papers lies at the heart of scientific advancement, but it requires substantial expertise, time, and effort. The peer review system faces several challenges, including a growing number of submissions that outpace the reviewer availability, lack of incentives, and reviewer fatigue~\citep{tropini2023time, horta2024crisis, hossain2025llmsmetareviewersassistantscase}. Large Language Models (LLMs) hold the potential to assist the peer review process by  automatically reviewing papers~\citep{Hosseini2023FightingRF, Robertson2023GPT4IS}, but can we trust LLM-generated reviews? Evaluating the quality of reviews is inherently complex due to their multi-dimensional nature. Researchers have employed various metrics for the evaluation such as surface-level (e.g. linguistic similarity to human reviews), content-level (e.g., relevance, specificity, and factual accuracy), and decision-level (e.g., accept/reject classification accuracy) metrics~\citep{ramachandran2017automated, du2024llms, liang2024can, zhou2024llm}.

% what is at stake if LLMs fail to focus correctly: trust, alignment, and blind spots - making trustworthy and expert-aligned reviews that even can reveal human's blind spots. 
However, existing evaluations fail to assess whether LLM reviews comprehensively address critical dimensions of papers. Evaluating the \textit{focus} of reviews is crucial because reviews with poor focus can negatively impact reviewers, even if they are accurate, relevant, and specific. For example, reviews that overly concentrate on methodological details while completely neglecting the novelty aspect of the proposed method could fail to suggest meaningful feedback, diverging from how expert reviewers assess the submission. It could also mislead junior reviewers by promoting incomplete perspectives and reinforce shallow paper review practices. Despite such importance, few attempts have been made to systematically evaluate whether the focus of LLM reviews aligns with that of expert reviews. Conducting the \textbf{focus-level evaluation} of LLM reviews is useful to reveal the blind spots of LLM reviews along with their central focus, offering important insights into how human reviewers can most effectively leverage LLM reviews in the peer review process. Moreover, it provides a concrete foundation for guiding LLM training toward more balanced and expert-aligned review behavior.

We introduce a framework for focus-level evaluation of LLM reviews, which systematically analyzes where the reviews direct their praise and criticism based on facets considered important in peer review (Figure~\ref{fig:task}). Given an LLM, the framework computes a \textbf{focus distribution}, a normalized distribution of how frequently review points (e.g., a list of strengths and weaknesses) address predefined facets (e.g., problem, method, and experiments) by leveraging a paper review dataset. The focus distribution can be computed by an automatic annotator that assigns a facet for each review point, enabling a fully automatic evaluation. The interpretable nature of the focus distribution provides actionable insights by clearly revealing which facets LLMs tend to emphasize or overlook in comparison to human experts.

% Given a paper review that consists of a list of strengths and weaknesses, our framework computes the focus of strengths $F_s$ and weaknesses $F_w$. We represent the focus as a normalized vector $F = \{p_1, p_2, .., p_n\}$ where $p_i$ refers to the proportion of review points that discuss a dimension $d_i$ in a list of points (e.g., strengths and weaknesses). Computing the focus vector is done by an automatic annotator, enabling a fully automatic evaluation. By aggregating the focus vectors for reviews in a dataset, we can understand which dimensions LLMs emphasize or overlook in reviewing papers, compared to human experts.

\begin{figure*}[t]
    \centering
    % Replace the placeholder box below with your actual figure code:
    % \fbox{\rule{0pt}{2in} \rule{.9\textwidth}{0pt}}
    \includegraphics[width=\linewidth]{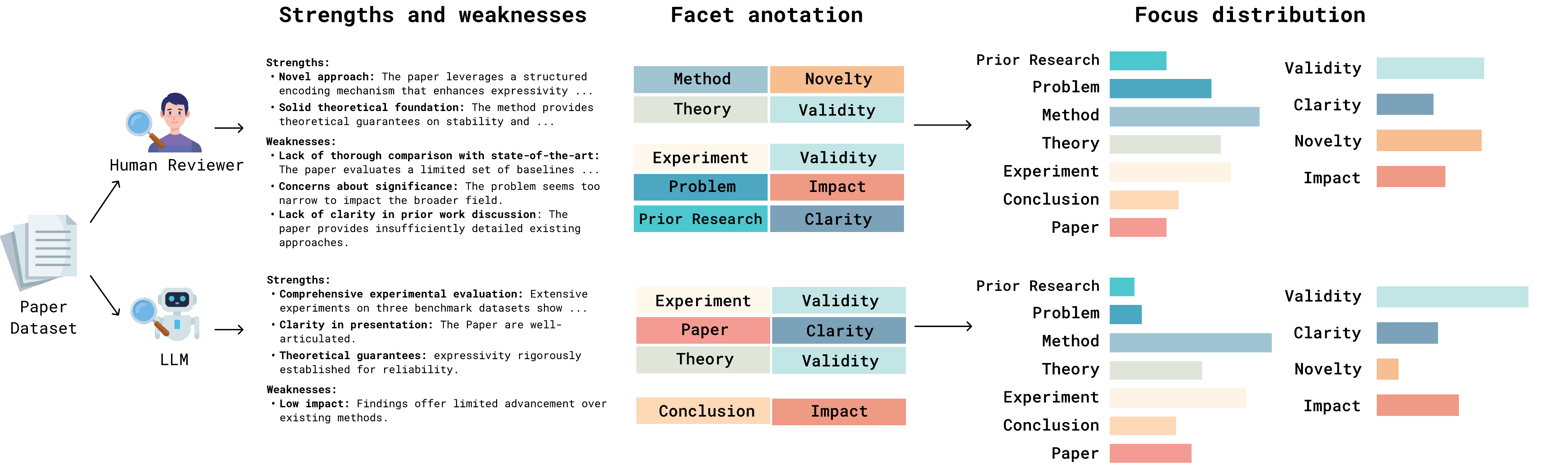}
    \caption{The overall process of automated focus-level evaluation. We first extracted strengths and weaknesses from review data on the OpenReview platform as the expert reviews. To identify key strengths and weaknesses influencing the final acceptance, we extracted them from the meta-review and augmented details from individual reviewer comments. Each strength and weakness was then annotated with a target and aspect by our automatic annotator. Finally, we computed the focus distributions by normalizing the frequency of annotated targets and aspects, and compare this distribution with that of LLM reviews.}
    \label{fig:review_generation}
\end{figure*}

To apply this framework for analyzing LLM-generated reviews in the context of AI conferences, we implemented a focus-level evaluation pipeline (Figure~\ref{fig:review_generation}). We identified the facets that constitute review focus, by surveying 9 paper submission guidelines from AI conferences and prior literature on review analysis~\citep{chakraborty2020aspect, ghosal2022peer, yuan2022can}.
% defining the facets that constitute the focus. 
We define two sets of facets: target (\textit{what} review points praise and critique such as problem, method, and experiment) and aspect (\textit{which criteria} is being evaluated such as validity, clarity, and novelty), which are key elements in analyzing paper reviews~\citep{ghosal2022peer, lu2025identifying}. We identified 7 facets for the target and 5 facets for the aspect (Table~\ref{tab:facets}). Next, we developed an automatic annotator for computing the focus distributions based on the target and aspect, which assigns a target and aspect label for a strength and weakness point in a review. The annotator showed substantial agreement with human annotators, achieving IRR (Cohen's kappa~\citep{cohen1960coefficient}) of 0.81 for target and 0.79 for aspect.

As a benchmark dataset for our focus-level evaluation pipeline, we constructed a dataset of 676 papers and their review data from OpenReview for ICLR conferences spanning 2021 to 2024. Then we computed and compared the focus distributions of human and LLM reviews using the evaluation pipeline (Figure~\ref{fig:radar_chart}), and we also measured text similarities between the reviews. Specifically, we evaluated 8 LLMs (4 GPT, 2 Llama, and 2 DeepSeek family) to analyze their review focus. We also evaluated MARG~\citep{d2024marg} as a novel review generation technique and a fine-tuned \texttt{gpt-4o} using our dataset. The results showed that:

\begin{itemize}
\item LLMs struggle to identify key targets and aspects in their reviews. Even the top-performing model reached an F1 score of 0.373 when matching human reviewers on the targets and aspects in each review point.
\item LLMs' review focus was biased towards examining technical validity, \textit{consistently overlooking novelty assessment} in weaknesses -- a critical limitation in paper review.
\item The fine-tuned model produced focus distributions most closely aligned with that of humans, compared to models using prompting alone.
\item The models demonstrated strengths in distinct areas. While the fine-tuned model produced the closest focus distributions, \texttt{Llama-405B} achieved the highest text similarity. It highlights the importance of holistic evaluation to capture the diverse aspects of review quality.

\end{itemize}

% Finally, we examine the agreement between LLMs and human experts (Figure~\ref{fig:task}) on the targets and aspects in the strengths and weaknesses. Specifically, we augmented meta-review, which is a validated review text from qualified experts, for identifying strengths and weaknesses identified by human experts (Figure~\ref{fig:augmentedReview}-a). Then we compared the targets and aspects of the strengths and weaknesses with these from LLM-generated reviews (Figure~\ref{fig:augmentedReview}-b). By introducing a systematic framework for examining different view points between LLMs and human experts in academic review, this work offers valuable insights into improving their performance and enhancing their potential role in assisting the review process.

We release a dataset comprising 676 papers, expert reviews, 3,657 strengths and weaknesses identified from the expert reviews with automatically annotated targets and aspects, LLM-generated reviews from 8 LLMs, and a total of 43,042 strengths and weaknesses extracted from the LLMs, each annotated with corresponding targets and aspects.

\section{A Framework for Focus-Level Evaluation of LLM Reviews}

We propose a \textit{focus-level evaluation} framework to systematically analyze what aspects LLMs emphasize or overlook when reviewing scientific papers. 
To enable interpretable and automated assessments of LLM behavior in reviewing, we aim to reveal the distribution of attention an LLM allocates to different review facets when identifying strengths and weaknesses in submissions. Specifically, we define a focus of the review to be compared for focus-level evaluation as follows: 

Let (i) $L$ be an LLM, (ii) $A=\{a_1, a_2, ..., a_N\}$ be a list of facets where each facet denotes a distinct criteria (e.g., problem, method, and experiment), and (iii) $P=\{p_1, p_2, ..., p_M\}$ be a corpus of paper submissions. The focus-level evaluation $E(L, A, P)$ produces two focus distributions $F^+$ and $F^-$ where $F^+$ denotes the distribution when identifying strengths of the submissions and $F^-$ for weaknesses. The focus distribution $F=(f_1, f_2, ..., f_N)$ can be represented as a normalized vector where $f_i$ denotes the relative frequency of review points (i.e., strengths for $F^+$ and weaknesses for $F^-$) that discuss the facet $a_i$, when $L$ generates reviews for paper submissions in $P$. 

% The focus distributions represent a relative attention of a language model when praising and criticizing papers, providing important insights into how to leverage the language model in the peer review process. 
To assess LLM behavior, our framework compares focus distributions with those from human expert reviews.
Researchers can specify the set of facets $A$ and the paper corpus $P$ based on the goals of their analysis, allowing flexible and targeted focus-evaluation.
% Computing the focus distributions can be automatically done by building an automatic annotator that assigns a facet $a$ for a single review point (i.e., a strength or a weakness), which enables a fully automated evaluation pipeline. 

Based on this framework, we implement an automatic focus-level evaluation pipeline to understand LLM's behavior in reviewing AI papers. Figure~\ref{fig:review_generation} illustrates the process of our focus-level evaluation pipeline. Our approach consists of three steps. (i) Collect an expert review dataset from ICLR conferences and extract strengths and weaknesses of the submissions for computing focus distributions of human experts (Section~\ref{sec: pipeline}), (ii) Define facets based on paper submission guidelines of AI conferences and build an automatic annotator based on the facets (Section~\ref{sec:automatic_annotator}), and (iii) Compute and analyze the focus distributions of LLMs and human experts in reviewing AI papers (Section~\ref{sec:evaluation}). 

% We define the paper review generation task as follows: given a research paper, 1) summarize main points, 2) identify a list of strengths and weaknesses, and 3) predict the final acceptance of the paper. This task offers direct value to various user groups (e.g., authors who want to get initial feedback on their draft, or reviewers who want to examine diverse viewpoints) by providing actionable feedback for improving their papers. While valuable, evaluating papers based on research standards (e.g., novelty, rigor, and clarity) is difficult for LLMs as it requires significant expertise. % Our evaluation does not consider the summary and the paper acceptance, as the acceptance decision depends on complex social factors beyond the given paper.

\section {Constructing Expert Review Dataset}
\label{sec: pipeline}

The focus-level evaluation framework requires a corpus of paper submissions $P$. We collected the review data from OpenReview platform and extracted the strengths and weaknesses of papers for computing focus distributions of human experts.

\subsection {Collecting Review Data}

We used real-world review data covering ICLR 2021-2024 from the OpenReview platform\footnote{The review data is publicly available and permits use of data for research.}, where human experts evaluated submissions for a top-tier AI conference. Using the OpenReview API\footnote{https://docs.openreview.net/getting-started/using-the-api} and the list of submissions from public GitHub repositories\footnote{https://github.com/\{evanzd/ICLR2021-OpenReviewData, fedebotu/ICLR2022-OpenReviewData, fedebotu/ICLR2023-OpenReviewData, hughplay/ICLR2024-OpenReviewData\} }, we initially collected 18,407 submissions with their review data. 

\subsection {Extracting Strengths and Weaknesses}

One of the challenges in identifying the strengths and weaknesses of these papers is that each review consists of multiple blocks, including a meta-review and individual reviews from several reviewers. To address the challenge, our approach is to use a meta-review, a final review from a qualified expert that summarizes reviews and highlights important strengths and weaknesses for supporting the final decision. As the meta-review does not capture all the details, we created self-contained strengths and weaknesses by 1) extracting them from the meta-review and 2) augmenting these extracted elements with detailed comments from individual reviews (non-meta). We designed a prompting chain that consists of three prompts (Appendix~\ref{appendix:gt-prompt}). % After excluding withdrawn submissions that lack meta-reviews, 14,922 submissions remained.

\section {Developing an Automatic Focus-level Evaluation Method}
\label{sec:automatic_annotator}

% The focus-level evaluation framework requires a list of facets $A$ along with an automatic annotator for enabling a fully automated evaluation. We analyzed AI paper submission guidelines to define facets for analyzing paper reviews. Specifically, we defined two sets of facets: \textbf{target} (\textit{what} the review praises or critiques) and \textbf{aspect} (\textit{which criteria} is being evaluated), which are key elements in analyzing paper reviews~\citep{ghosal2022peer, lu2025identifying}. Next, we developed an automatic annotator for assigning a target and aspect label, which achieved substantial agreement with human annotators.

To enable a fully automated evaluation using the proposed focus-level evaluation framework, we first define a set of facets and then develop an automatic annotator. We then compute focus distributions based on the annotated facets to analyze how LLMs and human reviewers differ in their focus of reviewing.

% The central goal of this paper is to analyze where LLMs excel and fall short in reviewing papers, compared to human experts. To achieve the goal, we 1) annotate each of the strengths and weaknesses identified by LLMs and experts and 2) examine the agreement between them based on the annotation results. The analysis offers insights into the distinct contributions and limitations of LLMs in reviewing papers, informing strategies to foster more effective human-LLM collaboration in reviewing papers.

\subsection{Defining Facets from Guidelines}

To build an initial set of facets, we surveyed 9 AI paper submission guidelines (Appendix~\ref{appendix:guidelines}) and extracted target-aspect pairs from each statement in the guidelines (e.g., \textit{``The paper should state the full set of assumptions of all theoretical results if the paper includes theoretical results.''} yields the target \textit{Theory} and aspect \textit{Completeness}). To ensure comprehensive coverage of facets, we also reviewed literature that analyzes paper review data~\citep{chakraborty2020aspect, ghosal2022peer, yuan2022can}. After identifying 33 targets and 13 aspects, we merged similar items to create simple and distinct categories, resulting in 7 targets and 4 aspects (Table~\ref{tab:facets}). The definition of each target and aspect facet is available in Appendix~\ref{appendix:target_and_aspect}.

\begin{table}[ht]
    \centering
    \small
    \setlength\tabcolsep{10pt}      % a bit of breathing room
    \renewcommand{\arraystretch}{1.1}
    \begin{tabular}{ll}
        \toprule
        \textbf{Target}      & \textbf{Aspect} \\ \midrule
        Problem              & Impact \\
        Prior Research       & Novelty \\
        Method               & Clarity \\
        Theory               & Validity \\
        Experiment           & Not\mbox{-}specific \\
        Conclusion           & \\        % (no matching aspect)
        Paper                & \\        % (no matching aspect)
        \bottomrule
    \end{tabular}
    \caption{Our research focuses on two sets of facets: target and aspect. Detailed definitions of the facets are available in Appendix~\ref{appendix:target_and_aspect}.}
    \label{tab:facets}
\end{table}

\subsection{Building Automatic Annotators}

Based on the identified facets, we annotated targets and aspects of strengths and weaknesses to produce ground truth for developing an automatic annotator. We randomly sampled 68 papers from our review dataset, yielding 327 instances of strengths and weaknesses. Two authors — one author is experienced in qualitative research in HCI and the other author has prior publications in the field of AI/NLP — synchronously decided each label together, resolving any conflicts. Most conflicts arose when an instance illustrated multiple points. For example, an instance such as ``\textit{**Technically sound with a strong foundation**: The paper's technical foundation is evident ... Technical novelty also arises from using supermartingale constraints ...}'' could correspond to both \textit{Validity} and \textit{Novelty} aspect. Two authors finalized the annotation through discussions, focusing on the main point or root cause of the issue. In the example, we annotated \textit{Validity}, as the strength mainly praises the technical soundness, as shown in the header wrapped in ``**''.

\begin{table}[ht]
    \centering
    \small
    \begin{tabular}{lcc}
        \toprule
        \textbf{Model}       & \textbf{Target} & \textbf{Aspect} \\
        \midrule
        \texttt{gpt-4o-mini} & 0.69   & 0.71   \\
        \texttt{gpt-4o}      & 0.83   & 0.75   \\
        \texttt{\textbf{o3-mini}}     & 0.81   & 0.79   \\
        \bottomrule
    \end{tabular}
    \caption{Inter-Rater Reliability (Cohen's kappa~\cite{cohen1960coefficient}) between annotations of authors and LLMs.}
    \label{tab:model_scores}
\end{table}

We then designed prompts to automatically annotate the instances, assigning a target and aspect label to each. Specifically, we designed four prompts where each corresponds to one of the four combinations of target/aspect and strength/weakness~\ref{appendix:annotate-prompt}. Table~\ref{tab:model_scores} shows the Inter-Rater Reliability (IRR, Cohen's kappa~\cite{cohen1960coefficient}) between human and LLM annotations for three language models. Annotation using \texttt{o3-mini} achieved the IRR scores of 0.81 for targets and 0.79 for aspects, indicating substantial agreement~\cite{cohen1960coefficient}. Given the high IRR and its relatively low computational cost compared to other two models, we used \texttt{o3-mini} for the automatic annotation of both target and aspect in the main evaluation. Moreover, an examination of the confusion matrix (Appendix~\ref{appendix:annotate-confusion-matrix}) suggests that the errors tend to occur in semantically related categories, indicating that the misclassifications are not arbitrary but rather reflect subtle ambiguities inherent in the data.

\subsection{Computing Focus Distributions}

Building on the defined facets and the automatic annotation method, we assign a target and aspect label to each strength and weakness point, using the automatic annotator. We then compute the normalized frequency of these labels to derive focus distributions of targets and aspects, respectively. Separate distributions are calculated for strengths and weaknesses, resulting in four distinct focus distributions. These focus distributions illustrate how LLMs and human reviewers allocate their attention across the different facets of a paper.

\section{Evaluation}
\label{sec:evaluation}
\begin{table*}[t]
    \centering
    \resizebox{\textwidth}{!}{%
    \begin{tabular}{l cccc ccc}
        \toprule
                     &
                     \multicolumn{4}{c}{\textbf{Focus similarity}} &
                     \multicolumn{3}{c}{\textbf{Text similarity}} \\
        \cmidrule(lr){2-5}\cmidrule(lr){6-8}
        Model & KL Divergence & Overall F1 & Strength F1 & Weakness F1
              & ROUGE-L & BERTScore & BLEU-4 \\ \midrule
        % ---------- baseline models ----------
        \texttt{gpt-4o-mini}   & \second{0.081} & 0.344 & 0.335 & 0.353 & 0.197 & \third{0.883} & 0.076 \\
        \texttt{gpt-4o}        & \third{0.082} & 0.348 & \second{0.342} & 0.354 & \third{0.202} & \first{0.885} & \third{0.079} \\
        \texttt{o1-mini}       & 0.090 & \second{0.359} & 0.331 & \third{0.385} & 0.179 & 0.878 & 0.059 \\
        \texttt{o1}            & 0.097 & \third{0.355} & 0.318 & \second{0.388} & 0.170 & 0.869 & 0.032 \\
        \texttt{DeepSeek-R1}   & 0.120 & \first{0.373} & \third{0.341} & \first{0.400} & 0.156 & 0.874 & 0.045 \\
        \texttt{Llama-70B}     & 0.136 & 0.339 & 0.338 & 0.341 & \second{0.215} & 0.882 & 0.076 \\
        \texttt{Llama-405B}    & 0.145 & 0.349 & \first{0.349} & 0.350 & \first{0.218} & \second{0.884} & \first{0.089} \\
        \texttt{DeepSeek-V3}   & 0.151 & 0.350 & 0.330 & 0.368 & 0.199 & 0.880 & 0.069 \\ \midrule
        % ---------- other techniques ----------
        \texttt{gpt-4o (FT)}   & \first{0.022} & 0.306 & 0.280 & 0.322 & 0.194 & 0.882 & \second{0.081} \\
        \texttt{MARG}          & 0.113 & 0.346 & \multicolumn{1}{c}{--} & 0.346 & 0.160 & 0.854 & 0.011 \\ \bottomrule
    \end{tabular}%
    }
    \caption{Overall performance by comparing expert reviews and LLM reviews. For focus similarity, we computed an average of the KL divergences of four focus distributions (strength/target, weakness/target, strength/aspect, and weakness/aspect) between LLM and expert reviews. The overall, strength, and weakness F1 scores were computed by comparing the (target, aspect) set between expert and LLM reviews. The text similariy metrics were computed between LLM reviews and expert reviews. The results highlight different areas of excellence across models (\texttt{gpt-4o (FT)}: the highest focus distribution similarity, \texttt{DeepSeek-R1}: the best agreement on (target, aspect) labels, \texttt{Llama-405B}: the highest text similarity score.)}
    \label{tab:metrics}
\end{table*}

\subsection{Setup}

\paragraph{Data.} The evaluation is based on paper-review pairs. However, we excluded \textit{accepted} submissions in the evaluation because OpenReview provides the camera-ready versions (post-review) rather than the submitted versions (pre-review), leading to a mismatch between the collected review and the camera-ready paper. Therefore, we only focused on \textit{rejected} papers, where the meta-review corresponds to the latest version of the paper. Out of 9,139 rejected papers, we randomly sampled 7.5\% of them (685 papers) for the evaluation. In total, we obtained 3,689 review items (1,241 strengths and 2,448 weaknesses), each automatically annotated with a target and aspect label.

For \textit{accepted} papers, we manually collected the submitted versions of a small sample (40 papers), which has the timestamp near the ICLR deadline in the version history in arXiv. See Appendix~\ref{appendix:result_accepted_paper} for the focus distribution results.

\begin{figure}[t]
  \centering
  \includegraphics[width=\linewidth]{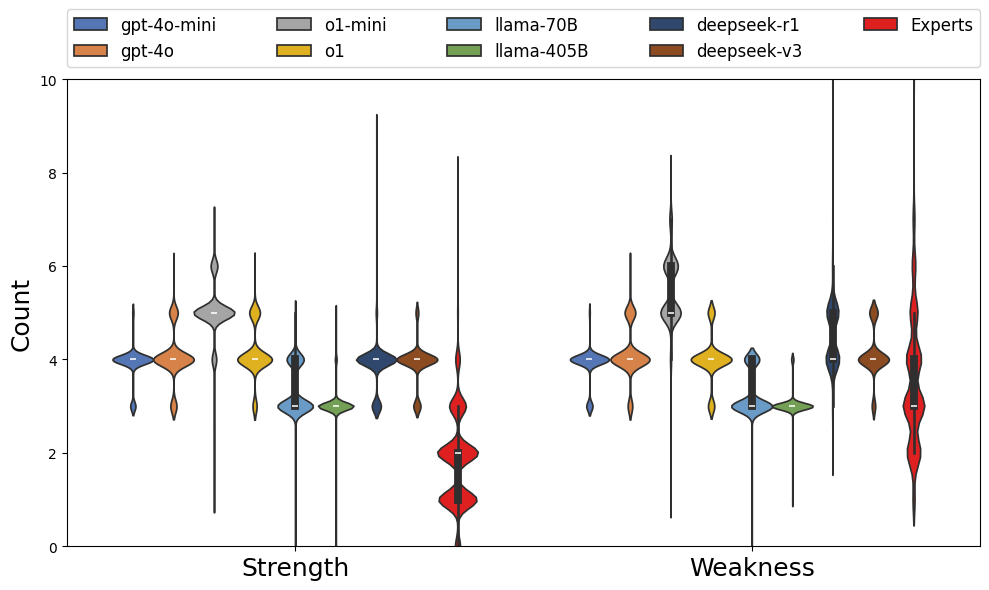}
  \caption{Distribution of strengths and weaknesses. Unlike human experts, LLMs reported a consistent count regardless of paper contents. \texttt{o1-mini} identified the most, while \texttt{Llama} models identified the fewest points.}
  \label{fig:num_reviews}
\end{figure}

\paragraph{Models.}
\label{sec:llm}We consider eight off-the-shelf LLMs, differing in size and availability (open-source vs. proprietary): four GPT models (gpt-4o-mini, gpt-4o, o1-mini, o3-mini, o1)\footnote{\texttt{gpt-4o-2024-08-06}, \texttt{gpt-4o-mini-2024-07-18}, \texttt{o1-mini-2024-09-12}, \texttt{o1-2024-12-17}}, two Llama models (Llama 3.1-\{70B, 405B\}), and two DeepSeek models (DeepSeek-\{V3, R1\}). We also evaluated MARG~\cite{d2024marg} and a fine-tuned \texttt{gpt-4o} (see Appendix~\ref{appendix:finetune} for the detail). For MARG, we only report scores for weaknesses because it only generates critiques of papers.

\paragraph{Metrics. } We employed two types of metrics: focus similarity and text similarity, used in prior work~\citep{zhou2024llm, chamoun2024automated, Gao2025ReviewAgentsBT}. For focus similarity, We measured Kullback-Leibler (KL) Divergence between the focus distributions of the models and human experts. We also measured F1 scores over the set of annotated $(\text{target}, \text{aspect})$ pairs as an agreement on review points. For text similarity, we measured ROUGE-L, BERTScore, and BLEU-4 between the LLM and expert reviews.

% Additionally, we report the number of points raised (i.e., strengths and weaknesses) in each review as a metric of review

% \subsection{Procedure}

% For each of the 685 papers, we computed focus distributions of human experts by extracting the strengths and weaknesses by human experts (Section~\ref{sec: pipeline}) and annotating the corresponding targets and aspects using the automatic annotator powered by \texttt{o3-mini}. Then for each LLM in Section~\ref{sec:llm}, we computed their focus distributions by generating a review for each paper (See Appendix~\ref{appendix:llm-prompt} for the prompt), extracting strengths and weaknesses in the review, and annotating the targets and aspects using the same automatic annotator. Then we compared the focus distributions between human experts and LLMs.

% \vspace{2mm}

\subsection{Result}

\begin{figure*}[t]
    \centering
    \includegraphics[width=\textwidth]{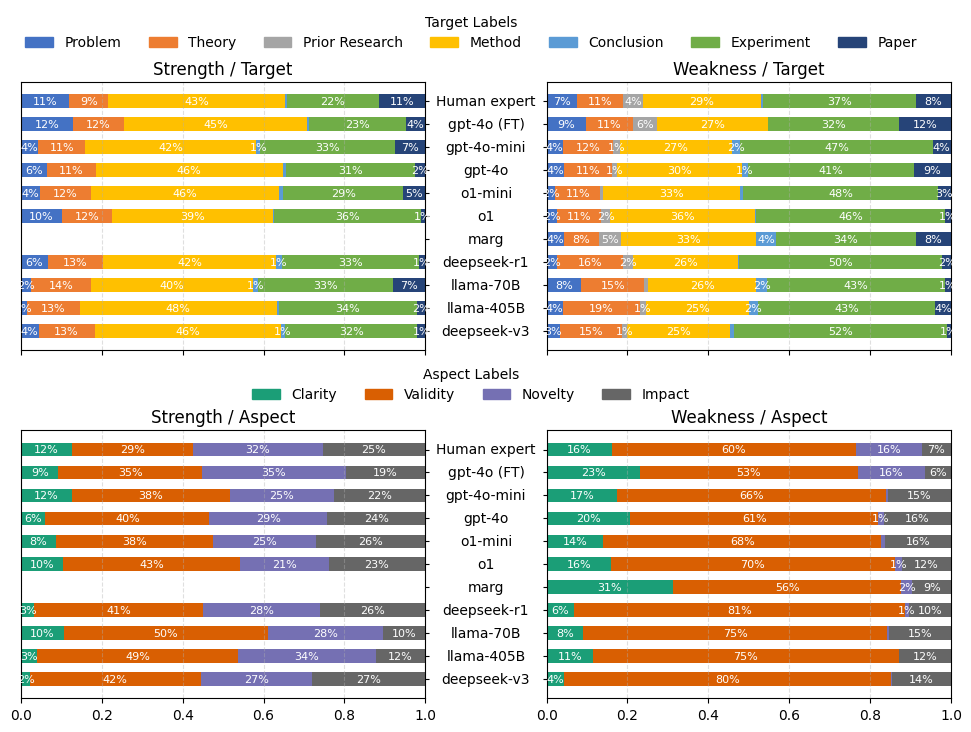}
    \caption{A visualization of focus distributions by target/aspect and strength/weakness, in a descending order of cosine similarity. Overall, both groups showed similar view points in reviewing papers, focusing on technical targets (i.e., Method, Experiment, and Theory) and validity. However, LLMs showed a more biased focus towards the technical validity whereas human experts exhibited more balanced focus. Moreover, all the LLMs lack consideration of Novelty for weaknesses compared to human experts, which is a significant limitation in reviewing papers.}
    \label{fig:radar_chart}
\end{figure*}

While human experts raised various number of points, LLMs identified a relatively consistent number of points regardless of the paper's content. Moreover, LLMs identified a similar number of points between strengths and weaknesses, which was a different pattern from that of the human experts (Figure~\ref{fig:num_reviews}). Overall, LLMs identified more points on average (7.88) than human experts (5.39). Among the LLMs, \texttt{Llama} models identified fewer (3.17 strengths and 3.15 weaknesses, on average) whereas \texttt{o1-mini} reported more strengths and weaknesses (5.03 and 5.47, respectively) than other models. The average review length of human experts and the models were 2639.76 and 3976.25, respectively. By comparing their focus distributions, we report the following key findings.

\textbf{The fine-tuned \texttt{gpt-4o} produced focus distributions most closely aligned with that of human experts, while other models excelled in different evaluation dimensions.} Table~\ref{tab:metrics} shows the overall performance of the models. \texttt{gpt-4o (FT)} showed the highest focus distribution similarity, \texttt{DeepSeek-R1} achieved the best agreement on (target, aspect) labels, and \texttt{Llama-405B} showed the highest text similarity score. \texttt{gpt-4o} showed balanced performance, with moderate scores for both focus and text similarity. The results indicate the multifaceted nature of the paper review evaluation task. In other words, assessing the quality of reviews needs a holistic approach that integrates multiple and complementary metrics.

% The precision and recall were computed by computing true positive, false positive, and false negative as \( TP = \sum \left| E_{r,c} \cap L_{r,c} \right|\), \( FP = \sum \left| L_{r,c} \setminus E_{r,c} \right|\), \(FN = \sum \left| E_{r,c} \setminus L_{r,c} \right| \) for \( r \in R, c \in \{\text{strength},\,\text{weakness}\} \), where $R$ is the set of papers, $c$ is the type of review item, and $E_{r, c}$ is the set of (target, aspect) that experts identified for the paper $r$ and the type $c$, and $L_{r, c}$ is the LLM-identified set.

\textbf{Overall, LLMs do not effectively identify key targets and aspects when reviewing papers.} Table~\ref{tab:metrics} shows the overall focus similarity and text similarity. The highest overall F1 score among the LLMs was 0.373, which indicates a low level of agreement with human experts in identifying strengths and weaknesses. Since we only considered whether the categories of review items match rather than their detailed content, the result implies that the actual content of strengths and weaknesses is significantly different between human experts and LLMs. In general, LLMs showed higher recall (0.402) than precision (0.300) scores, mainly due to the nature of identifying a higher number of review points than human experts. Also, LLMs consistently achieved higher F1 scores for weaknesses than strengths.

\textbf{While overall agreement is low, both groups have similar primary focus in reviewing papers.} Figure~\ref{fig:radar_chart} shows a visualization of focus distributions between LLMs and human experts. For targets, both groups primarily focused on core technical elements---Method, Experiment, and Theory. However, strengths and weaknesses illustrated different patterns: both groups praised Method more than Experiment in the strengths, but criticized Experiment more than Method in the weaknesses. For aspects, both groups considered Validity as the primary focus when identifying weaknesses. However, human experts focused more on Novelty in strengths whereas LLMs maintained Validity as the primary focus. For both groups, Impact received more attention in the strengths than weaknesses, whereas Clarity showed the opposite.

 % While Method and Experiment showed the high level of agreement as both groups share the core perspective, Problem and Theory showed low agreement. For Theory, F1-score was consistently higher for weaknesses than strengths, suggesting that LLMs are more effective at identifying concerns in theories (e.g., the assumptions are too strong) than recognizing strong points of theories (e.g., the theoretical analysis is thorough). For aspects, Validity showed the highest agreement, but it would be due to our focus of \textit{rejected} papers. Specifically, we observed that LLMs almost always report Validity concerns in weaknesses (Recall $\geq$ 0.99 on average) where rejected papers might contain such limitations. Novelty aspect showed low agreement, particularly for weaknesses, which is due to the lack of focus in Novelty in evaluating weaknesses.

\textbf{LLMs \textit{consistently} exhibited a more biased focus, notably overlooking \textit{novelty assessment} in identifying weaknesses.} Although both groups had the similar primary focus, LLMs tend to concentrate on a few specific dimensions. For instance, for targets, LLMs focused primarily on Method and Experiment, with less focus on Prior Research (e.g., whether the paper adequately addresses prior work in positioning) and Problem (e.g., whether the task needs community attention) compared to human experts (Problem in the strengths and Prior Research in the weaknesses). For aspects, LLMs mostly focused on Validity in both strengths and weaknesses. In contrast, human experts considered the aspects more evenly. The LLMs' biased focus was observed for \textit{accepted} papers too, mostly criticizing experimental validity (See Appendix~\ref{appendix:result_accepted_paper}). Notably, LLMs rarely focused on Novelty aspect in identifying weaknesses. This is a significant drawback, as a paper review requires a critical examination of novelty, by comparing them against existing work. Fortunately, we observed that \texttt{gpt-4o (FT)} identifies Novelty aspect in the weakness, as close as human experts.

Due to their biased focus, the level of agreement between LLMs and human experts varied across different labels. For targets and aspects that LLMs primarily focus on --- Method (0.731, an average F1 score) and Experiment (0.671) targets and Validity (0.771) aspect --- LLMs had a much higher level of agreement with human experts compared to other targets (0.213) and aspects (0.340). In the case of Experiment, the F1 score was consistently higher for weaknesses (0.835) than strengths (0.513), suggesting that LLMs are more effective at identifying concerns (e.g., lack of baselines or scope of evaluation) than strong points of experiments (e.g., experiments are rigorous and thorough). Similarly, for aspects other than Validity, agreement levels were notably lower. In particular, Novelty in the weaknesses, which LLMs largely overlooked, showed a significantly lower F1 score (0.126). See Appendix~\ref{appendix:result} for the full results.

% Due to their biased focus, the level of agreement between LLMs and human experts varied across different labels. Table~\ref{tab:merged_sw} shows F1 scores for specific targets and aspects. For targets and aspects that LLMs focus more on --- Method and Experiment targets and Validity aspect --- LLMs had a much higher level of agreement with human experts compared to other targets and aspects. In the case of Experiment, the F1 score was consistently higher for weaknesses than strengths, suggesting that LLMs are more effective at identifying concerns in experiments (e.g., lack of baselines or scope of evaluation) than recognizing strong points of experiments (e.g., experiments are rigorous and thorough). Similarly, for aspects other than Validity, agreement levels were notably lower. In particular, Novelty in the weaknesses, which LLMs largely overlooked, showed a significantly lower F1 score.

% Validity showed the highest agreement, but it would be due to our focus of \textit{rejected} papers. Specifically, we observed that LLMs almost always report Validity concerns in weaknesses (Recall $\geq$ 0.99 on average) where rejected papers might contain such limitations. Novelty aspect showed low agreement, particularly for weaknesses, which is due to the lack of focus in Novelty in evaluating weaknesses.

\textbf{LLMs showed similar patterns in their focus, regardless of their size and reasoning capability.} All LLMs, including both proprietary and open source models, showed similar patterns that focused primarily on technical (Method, Experiment, and Theory) validity than on Novelty for the weaknesses. This consistency indicates that the observed biases could stem from the inherent design and training methods of LLMs, revealing potential room for improvement in the reasoning capability that requires leveraging external information (e.g., identifying comparable related work and analyzing novelty of submissions).

\section{Discussion}

In this paper, we found gaps between human experts and LLMs about their focus in reviewing papers and reported several limitations of LLMs as an automated reviewer. Based on the results, we discuss the following implications.

% \newpage
\textbf{There is significant room for improving alignments between human experts and LLMs in paper reviewing.} Our results show that LLMs exhibit a more biased focus, primarily assessing technical validity without contextual consideration, compared to human experts. While fine-tuning yielded closer focus with human experts, the alignment of review points remained low. Since our focus-level evaluation only considered the target and aspect labels rather than their actual contents, we suspect that a more significant gap lies in the actual content addressed in the review items. For instance, even if two review points share the same label set (Experiment, Validity), they could point out different points such as lack of necessary baselines or lack of ablation studies to justify authors' arguments. Content-level investigations based on annotated facets may reveal more specific limitations of LLMs in reviewing papers, ultimately contributing to improving their reasoning capability.

% A checklist-based approach, which asks the LLM to generate strengths and weaknesses that match those of human experts, can be a promising direction.
 % Also, research could investigate how to design Human-AI interactions by leveraging what LLMs excel at. For instance, LLMs can be effective in rigorously examining technical validity. % We believe that our automatic evaluation pipeline can be used for evaluating the alignments, evolved by updating recent publication data. 
%  For LLM-powered paper reviews to provide value to various user groups (e.g., novice researchers seeking feedback or reviewers evaluating submissions), it is important to reduce the gap.  

\textbf{Focus-level evaluation reveals the complementary strengths of human reviewers and LLM reviewers.} Our evaluation shows that LLM reviews tend to emphasize technical validity, whereas human reviews offer a more balanced perspective. These differences motivate the design of a paper review pipeline that integrates the strengths of both human and LLM reviews. For instance, LLMs could be purposefully used to perform systematic validity checks that humans may overlook due to fatigue~\citep{tropini2023time, horta2024crisis, hossain2025llmsmetareviewersassistantscase}, while humans provide more nuanced judgements on novelty and significance. By examining review focus, we not only uncover blind spots in the review process but also generate concrete guidance for integrating human and LLM reviewers to improve the overall paper review process.

\textbf{Research should investigate the task of assessing the novelty of academic papers.} Our finding illustrated that all untuned LLMs in our analysis significantly overlooked the novelty aspect when evaluating weaknesses of papers. Previous studies have indicated that language models’ ability to assess novelty is inferior to that of experts~\cite{Just02072024, Lin2024EvaluatingAE}, emphasizing the need to encourage LLMs to focus on novelty evaluation. Although novelty is one of the most important aspects in reviewing papers and efforts have been made to enhance LLMs' ability to assess novelty~\citep{Bougie2024GenerativeAR, Lin2024EvaluatingAE}, there exists no suitable benchmark for systematically measuring their novelty assessment capability. We believe that creating the benchmark is a valuable contribution to the field, allowing LLMs to learn how to assess similarities between papers. Leveraging data in OpenReview could be an initial step as it contains experts' judgment on novelty of the paper for both positive and negative decisions.

\textbf{A focus-level evaluation framework can offer unique value for guiding LLM training.} The automated focus-level evaluation pipeline enables continuously tracking and evaluation of how LLMs focus on key facets of a paper over time, which aligns with the goals of holistic evaluation benchmarks~\citep{liang2022holistic, srivastava2022beyond}. Beyond the language model evaluation, focus-level supervision can be incorporated during the training process; reward functions can be designed to encourage balanced focus aligned with human experts or even purposefully facilitate a certain focus (e.g., building a novelty-focused reviewer)~\citep{yang2024rewards, agnihotri2025multiobjectivepreferenceoptimizationimproving}. Furthermore, the framework is generalizable to other domains where the output spans multiple facets---such as debating, decision making, and educational feedback---making \textit{focus} a critical factor in generated outputs. 
% is important during the reasoning process, including . 
% With the interpretable and controllable the reasoning process in a focus-level, our framework enables building more transparent, goal-aligned, and trustworthy language models.

\textbf{}

\section{Related Work}

With the powerful reasoning capability of LLMs, LLMs have the potential to assist in the task of reviewing papers~\citep{latona2024ai, d2024marg}. Research has explored the capability of LLMs in reviewing papers, identifying a set of limitations. While LLM-generated reviews can be helpful~\citep{liang2024can, tyser2024ai, Lu2024TheAS}, research has shown that LLMs-generated reviews lack diversity~\citep{du2024llms, liang2024can} and technical details~\citep{zhou2024llm}, exhibit bias~\citep{Ye2024AreWT}, tend to provide positive feedback~\citep{zhou2024llm, du2024llms}, and may include irrelevant or even inaccurate comments~\citep{mostafapour2024evaluating}. Furthermore, research also has reported that LLM-generated reviews have a low level of agreement with experts-generated reviews~\citep{saad2024exploring}. 

To assess the quality of review, research has taken a quantitative approach by analyzing review text. For instance, research has evaluated the quality of review based on human preferences~\citep{tyser2024ai}, similarity to human-generated review~\citep{zhou2024llm, liang2024can, gao2024reviewer2optimizingreviewgeneration, Sun2024MetaWriterET, chamoun2024automated} and classification-based scores~\citep{Li2023SummarizingMD}. Another approach is to classify review data based on categories such as section~\citep{ghosal2022peer}, aspect~\citep{yuan2022can, chamoun2024automated, liang2024can} and actionability~\citep{Choudhary2022ReActAR}. While quantitative approach provides concrete insights, it is typically conducted as a one-time evaluation, challenging to apply the consistent methodology over time.

\section{Conclusion}

We introduced a framework for focus-level evaluation of LLM reviews, which systematically analyzes where LLM reviews direct their praise and criticism based on pre-defined facets. Our findings suggest that LLMs need to adopt a more balanced perspective, have higher agreement with human experts about the target and aspect in the strengths and weaknesses, and place greater emphasis on novelty assessment when criticizing papers. We believe that the focus-level evaluation can contribute to ongoing evaluation of LLMs' paper review capabilities within the rapid pace of LLM developments.
% , offering concrete insights for improving their reasoning capability.

\section*{Acknowledgments}
This work was conducted with the support of LG AI Research, which is gratefully acknowledged.

\section*{Limitation}

This paper has the following limitations. First, our dataset focuses solely on ICLR submissions and the coding schema is developed based on AI venues, which limit generalizability to other fields. Second, our analysis examines the target and aspect of the review items, but other important dimensions such as level of specificity and depth of justification remain unexplored. Third, while our automatic annotator achieved high IRR (0.80) with human annotations, some discrepancies still exist. Finally, we did not explore possible prompt engineering strategies that could mitigate the limitations of LLMs in paper review. Future work can investigate techniques to enhance the alignment between LLMs and human experts.

\section*{Ethical impact}
This paper presents potential risks. First, while our vision is to build LLMs to effectively assist review process, our work could inadvertently encourage over-reliance on LLM-generated reviews among various user groups, including reviewers and novice researchers. Second, although our dataset could contribute to improving LLM performance of reviewing papers, it may introduce a certain bias due to the source of dataset; ICLR for papers and code based on AI research. Finally, we assess the quality of review based on alignment with expert reviews, but it could offer a potentially biased perspective, as our facets only considers two dimensions, which may undervalue the unique contributions of LLM-generated reviews.

\bibliography{main}

\begin{thebibliography}{34}
\providecommand{\natexlab}[1]{#1}

\bibitem[{Agnihotri et~al.(2025)Agnihotri, Jain, Ramachandran, and Wen}]{agnihotri2025multiobjectivepreferenceoptimizationimproving}
Akhil Agnihotri, Rahul Jain, Deepak Ramachandran, and Zheng Wen. 2025.
\newblock \href {https://arxiv.org/abs/2505.10892} {Multi-objective preference optimization: Improving human alignment of generative models}.
\newblock \emph{Preprint}, arXiv:2505.10892.

\bibitem[{Bougie and Watanabe(2024)}]{Bougie2024GenerativeAR}
Nicolas Bougie and Narimasa Watanabe. 2024.
\newblock \href {https://api.semanticscholar.org/CorpusID:274776902} {Generative adversarial reviews: When llms become the critic}.
\newblock \emph{ArXiv}, abs/2412.10415.

\bibitem[{Chakraborty et~al.(2020)Chakraborty, Goyal, and Mukherjee}]{chakraborty2020aspect}
Souvic Chakraborty, Pawan Goyal, and Animesh Mukherjee. 2020.
\newblock Aspect-based sentiment analysis of scientific reviews.
\newblock In \emph{Proceedings of the ACM/IEEE Joint Conference on Digital Libraries in 2020}, pages 207--216.

\bibitem[{Chamoun et~al.(2024)Chamoun, Schlichktrull, and Vlachos}]{chamoun2024automated}
Eric Chamoun, Michael Schlichktrull, and Andreas Vlachos. 2024.
\newblock Automated focused feedback generation for scientific writing assistance.
\newblock \emph{arXiv preprint arXiv:2405.20477}.

\bibitem[{Choudhary et~al.(2022)Choudhary, Modani, and Maurya}]{Choudhary2022ReActAR}
G.~Choudhary, Natwar Modani, and Nitish Maurya. 2022.
\newblock \href {https://api.semanticscholar.org/CorpusID:244852407} {React: A review comment dataset for actionability (and more)}.
\newblock \emph{ArXiv}, abs/2210.00443.

\bibitem[{Cohen(1960)}]{cohen1960coefficient}
Jacob Cohen. 1960.
\newblock A coefficient of agreement for nominal scales.
\newblock \emph{Educational and psychological measurement}, 20(1):37--46.

\bibitem[{D'Arcy et~al.(2024)D'Arcy, Hope, Birnbaum, and Downey}]{d2024marg}
Mike D'Arcy, Tom Hope, Larry Birnbaum, and Doug Downey. 2024.
\newblock Marg: Multi-agent review generation for scientific papers.
\newblock \emph{arXiv preprint arXiv:2401.04259}.

\bibitem[{Du et~al.(2024)Du, Wang, Zhao, Deng, Liu, Lou, Zou, Venkit, Zhang, Srinath et~al.}]{du2024llms}
Jiangshu Du, Yibo Wang, Wenting Zhao, Zhongfen Deng, Shuaiqi Liu, Renze Lou, Henry~Peng Zou, Pranav~Narayanan Venkit, Nan Zhang, Mukund Srinath, et~al. 2024.
\newblock Llms assist nlp researchers: Critique paper (meta-) reviewing.
\newblock \emph{arXiv preprint arXiv:2406.16253}.

\bibitem[{Gao et~al.(2025)Gao, Ruan, Gao, Liu, and Fu}]{Gao2025ReviewAgentsBT}
Xian Gao, Jiacheng Ruan, Jingsheng Gao, Ting Liu, and Yuzhuo Fu. 2025.
\newblock \href {https://api.semanticscholar.org/CorpusID:276928587} {Reviewagents: Bridging the gap between human and ai-generated paper reviews}.
\newblock \emph{ArXiv}, abs/2503.08506.

\bibitem[{Gao et~al.(2024)Gao, Brantley, and Joachims}]{gao2024reviewer2optimizingreviewgeneration}
Zhaolin Gao, Kianté Brantley, and Thorsten Joachims. 2024.
\newblock \href {https://arxiv.org/abs/2402.10886} {Reviewer2: Optimizing review generation through prompt generation}.
\newblock \emph{Preprint}, arXiv:2402.10886.

\bibitem[{Ghosal et~al.(2022)Ghosal, Kumar, Bharti, and Ekbal}]{ghosal2022peer}
Tirthankar Ghosal, Sandeep Kumar, Prabhat~Kumar Bharti, and Asif Ekbal. 2022.
\newblock Peer review analyze: A novel benchmark resource for computational analysis of peer reviews.
\newblock \emph{Plos one}, 17(1):e0259238.

\bibitem[{Horta and Jung(2024)}]{horta2024crisis}
Hugo Horta and Jisun Jung. 2024.
\newblock The crisis of peer review: Part of the evolution of science.
\newblock \emph{Higher Education Quarterly}, page e12511.

\bibitem[{Hossain et~al.(2025)Hossain, Sinha, Bansal, Knipper, Sarkar, Salvador, Mahajan, Guttikonda, Akter, Hassan, Freestone, Jr., Feng, and Karmaker}]{hossain2025llmsmetareviewersassistantscase}
Eftekhar Hossain, Sanjeev~Kumar Sinha, Naman Bansal, Alex Knipper, Souvika Sarkar, John Salvador, Yash Mahajan, Sri Guttikonda, Mousumi Akter, Md.~Mahadi Hassan, Matthew Freestone, Matthew C.~Williams Jr., Dongji Feng, and Santu Karmaker. 2025.
\newblock \href {https://arxiv.org/abs/2402.15589} {Llms as meta-reviewers' assistants: A case study}.
\newblock \emph{Preprint}, arXiv:2402.15589.

\bibitem[{Hosseini and Horbach(2023)}]{Hosseini2023FightingRF}
Mohammad Hosseini and Serge~P.J.M. Horbach. 2023.
\newblock \href {https://api.semanticscholar.org/CorpusID:257310820} {Fighting reviewer fatigue or amplifying bias? considerations and recommendations for use of chatgpt and other large language models in scholarly peer review}.
\newblock \emph{Research Integrity and Peer Review}, 8.

\bibitem[{Julian~Just and Hutter(2024)}]{Just02072024}
Johann~Füller Julian~Just, Thomas~Ströhle and Katja Hutter. 2024.
\newblock \href {https://doi.org/10.1080/14479338.2023.2215740} {Ai-based novelty detection in crowdsourced idea spaces}.
\newblock \emph{Innovation}, 26(3):359--386.

\bibitem[{Latona et~al.(2024)Latona, Ribeiro, Davidson, Veselovsky, and West}]{latona2024ai}
Giuseppe~Russo Latona, Manoel~Horta Ribeiro, Tim~R Davidson, Veniamin Veselovsky, and Robert West. 2024.
\newblock The ai review lottery: Widespread ai-assisted peer reviews boost paper scores and acceptance rates.
\newblock \emph{arXiv preprint arXiv:2405.02150}.

\bibitem[{Li et~al.(2023)Li, Hovy, and Lau}]{Li2023SummarizingMD}
Miao Li, Eduard~H. Hovy, and Jey~Han Lau. 2023.
\newblock \href {https://api.semanticscholar.org/CorpusID:258436793} {Summarizing multiple documents with conversational structure for meta-review generation}.
\newblock In \emph{Conference on Empirical Methods in Natural Language Processing}.

\bibitem[{Liang et~al.(2022)Liang, Bommasani, Lee, Tsipras, Soylu, Yasunaga, Zhang, Narayanan, Wu, Kumar et~al.}]{liang2022holistic}
Percy Liang, Rishi Bommasani, Tony Lee, Dimitris Tsipras, Dilara Soylu, Michihiro Yasunaga, Yian Zhang, Deepak Narayanan, Yuhuai Wu, Ananya Kumar, et~al. 2022.
\newblock Holistic evaluation of language models.
\newblock \emph{arXiv preprint arXiv:2211.09110}.

\bibitem[{Liang et~al.(2024)Liang, Zhang, Cao, Wang, Ding, Yang, Vodrahalli, He, Smith, Yin et~al.}]{liang2024can}
Weixin Liang, Yuhui Zhang, Hancheng Cao, Binglu Wang, Daisy~Yi Ding, Xinyu Yang, Kailas Vodrahalli, Siyu He, Daniel~Scott Smith, Yian Yin, et~al. 2024.
\newblock Can large language models provide useful feedback on research papers? a large-scale empirical analysis.
\newblock \emph{NEJM AI}, 1(8):AIoa2400196.

\bibitem[{Lin et~al.(2024)Lin, Peng, and Fang}]{Lin2024EvaluatingAE}
Ethan Lin, Zhiyuan Peng, and Yi~Fang. 2024.
\newblock \href {https://api.semanticscholar.org/CorpusID:272881158} {Evaluating and enhancing large language models for novelty assessment in scholarly publications}.
\newblock \emph{ArXiv}, abs/2409.16605.

\bibitem[{Lu et~al.(2024)Lu, Lu, Lange, Foerster, Clune, and Ha}]{Lu2024TheAS}
Chris Lu, Cong Lu, Robert~Tjarko Lange, Jakob~N. Foerster, Jeff Clune, and David Ha. 2024.
\newblock \href {https://api.semanticscholar.org/CorpusID:271854887} {The ai scientist: Towards fully automated open-ended scientific discovery}.
\newblock \emph{ArXiv}, abs/2408.06292.

\bibitem[{Lu et~al.(2025)Lu, Kuznetsov, and Gurevych}]{lu2025identifying}
Sheng Lu, Ilia Kuznetsov, and Iryna Gurevych. 2025.
\newblock Identifying aspects in peer reviews.
\newblock \emph{arXiv preprint arXiv:2504.06910}.

\bibitem[{Mostafapour et~al.(2024)Mostafapour, Fortier, Pacheco, Murray, and Garber}]{mostafapour2024evaluating}
Mehrnaz Mostafapour, Jacqueline~H Fortier, Karen Pacheco, Heather Murray, and Gary Garber. 2024.
\newblock Evaluating literature reviews conducted by humans versus chatgpt: Comparative study.
\newblock \emph{Jmir ai}, 3:e56537.

\bibitem[{Ramachandran et~al.(2017)Ramachandran, Gehringer, and Yadav}]{ramachandran2017automated}
Lakshmi Ramachandran, Edward~F Gehringer, and Ravi~K Yadav. 2017.
\newblock Automated assessment of the quality of peer reviews using natural language processing techniques.
\newblock \emph{International Journal of Artificial Intelligence in Education}, 27(3):534--581.

\bibitem[{Robertson(2023)}]{Robertson2023GPT4IS}
Zachary Robertson. 2023.
\newblock \href {https://api.semanticscholar.org/CorpusID:259837446} {Gpt4 is slightly helpful for peer-review assistance: A pilot study}.
\newblock \emph{ArXiv}, abs/2307.05492.

\bibitem[{Saad et~al.(2024)Saad, Jenko, Ariyaratne, Birch, Iyengar, Davies, Vaishya, and Botchu}]{saad2024exploring}
Ahmed Saad, Nathan Jenko, Sisith Ariyaratne, Nick Birch, Karthikeyan~P Iyengar, Arthur~Mark Davies, Raju Vaishya, and Rajesh Botchu. 2024.
\newblock Exploring the potential of chatgpt in the peer review process: an observational study.
\newblock \emph{Diabetes \& Metabolic Syndrome: Clinical Research \& Reviews}, 18(2):102946.

\bibitem[{Srivastava et~al.(2022)Srivastava, Rastogi, Rao, Shoeb, Abid, Fisch, Brown, Santoro, Gupta, Garriga-Alonso et~al.}]{srivastava2022beyond}
Aarohi Srivastava, Abhinav Rastogi, Abhishek Rao, Abu Awal~Md Shoeb, Abubakar Abid, Adam Fisch, Adam~R Brown, Adam Santoro, Aditya Gupta, Adri{\`a} Garriga-Alonso, et~al. 2022.
\newblock Beyond the imitation game: Quantifying and extrapolating the capabilities of language models.
\newblock \emph{arXiv preprint arXiv:2206.04615}.

\bibitem[{Sun et~al.(2024)Sun, Tao, Hu, and Dow}]{Sun2024MetaWriterET}
Lu~Sun, Stone Tao, Junjie Hu, and Steven~P. Dow. 2024.
\newblock \href {https://api.semanticscholar.org/CorpusID:269470548} {Metawriter: Exploring the potential and perils of ai writing support in scientific peer review}.
\newblock \emph{Proceedings of the ACM on Human-Computer Interaction}, 8:1 -- 32.

\bibitem[{Tropini et~al.(2023)Tropini, Finlay, Nichter, Melby, Metcalf, Dominguez-Bello, Zhao, McFall-Ngai, Geva-Zatorsky, Amato et~al.}]{tropini2023time}
Carolina Tropini, B~Brett Finlay, Mark Nichter, Melissa~K Melby, Jessica~L Metcalf, Maria~Gloria Dominguez-Bello, Liping Zhao, Margaret~J McFall-Ngai, Naama Geva-Zatorsky, Katherine~R Amato, et~al. 2023.
\newblock Time to rethink academic publishing: the peer reviewer crisis.

\bibitem[{Tyser et~al.(2024)Tyser, Segev, Longhitano, Zhang, Meeks, Lee, Garg, Belsten, Shporer, Udell et~al.}]{tyser2024ai}
Keith Tyser, Ben Segev, Gaston Longhitano, Xin-Yu Zhang, Zachary Meeks, Jason Lee, Uday Garg, Nicholas Belsten, Avi Shporer, Madeleine Udell, et~al. 2024.
\newblock Ai-driven review systems: evaluating llms in scalable and bias-aware academic reviews.
\newblock \emph{arXiv preprint arXiv:2408.10365}.

\bibitem[{Yang et~al.(2024)Yang, Pan, Luo, Qiu, Zhong, Yu, and Chen}]{yang2024rewards}
Rui Yang, Xiaoman Pan, Feng Luo, Shuang Qiu, Han Zhong, Dong Yu, and Jianshu Chen. 2024.
\newblock Rewards-in-context: Multi-objective alignment of foundation models with dynamic preference adjustment.
\newblock \emph{arXiv preprint arXiv:2402.10207}.

\bibitem[{Ye et~al.(2024)Ye, Pang, Chai, Chen, fei Yin, Xiang, Dong, Shao, and Chen}]{Ye2024AreWT}
Rui Ye, Xianghe Pang, Jingyi Chai, Jiaao Chen, Zhen fei Yin, Zhen Xiang, Xiaowen Dong, Jing Shao, and Siheng Chen. 2024.
\newblock \href {https://api.semanticscholar.org/CorpusID:274436760} {Are we there yet? revealing the risks of utilizing large language models in scholarly peer review}.
\newblock \emph{ArXiv}, abs/2412.01708.

\bibitem[{Yuan et~al.(2022)Yuan, Liu, and Neubig}]{yuan2022can}
Weizhe Yuan, Pengfei Liu, and Graham Neubig. 2022.
\newblock Can we automate scientific reviewing?
\newblock \emph{Journal of Artificial Intelligence Research}, 75:171--212.

\bibitem[{Zhou et~al.(2024)Zhou, Chen, and Yu}]{zhou2024llm}
Ruiyang Zhou, Lu~Chen, and Kai Yu. 2024.
\newblock Is llm a reliable reviewer? a comprehensive evaluation of llm on automatic paper reviewing tasks.
\newblock In \emph{Proceedings of the 2024 Joint International Conference on Computational Linguistics, Language Resources and Evaluation (LREC-COLING 2024)}, pages 9340--9351.

\end{thebibliography}

\clearpage
\appendix
\begin{onecolumn}
\section{Appendix}

\label{sec:appendix}

\subsection{Review Generation}
\subsubsection{Prompts for Expert Review Generation}
\label{appendix:gt-prompt}

In this section, we provide prompts for identifying key strength and weakness from review data. Figure~\ref{fig:metareview-summarization} shows the prompt for extracting weakness and strength from meta-review. Figure~\ref{fig:augmentedReview} shows the prompt for using detailed comments from reviews to augment the extracted elements. Figure~\ref{fig:paraphrasing} shows the prompt for removing some extraneous reference. We used the three prompts in a prompt chain, sequentially running the prompts.

\begin{figure*}[h]
    \centering
    \includegraphics[width=\linewidth]{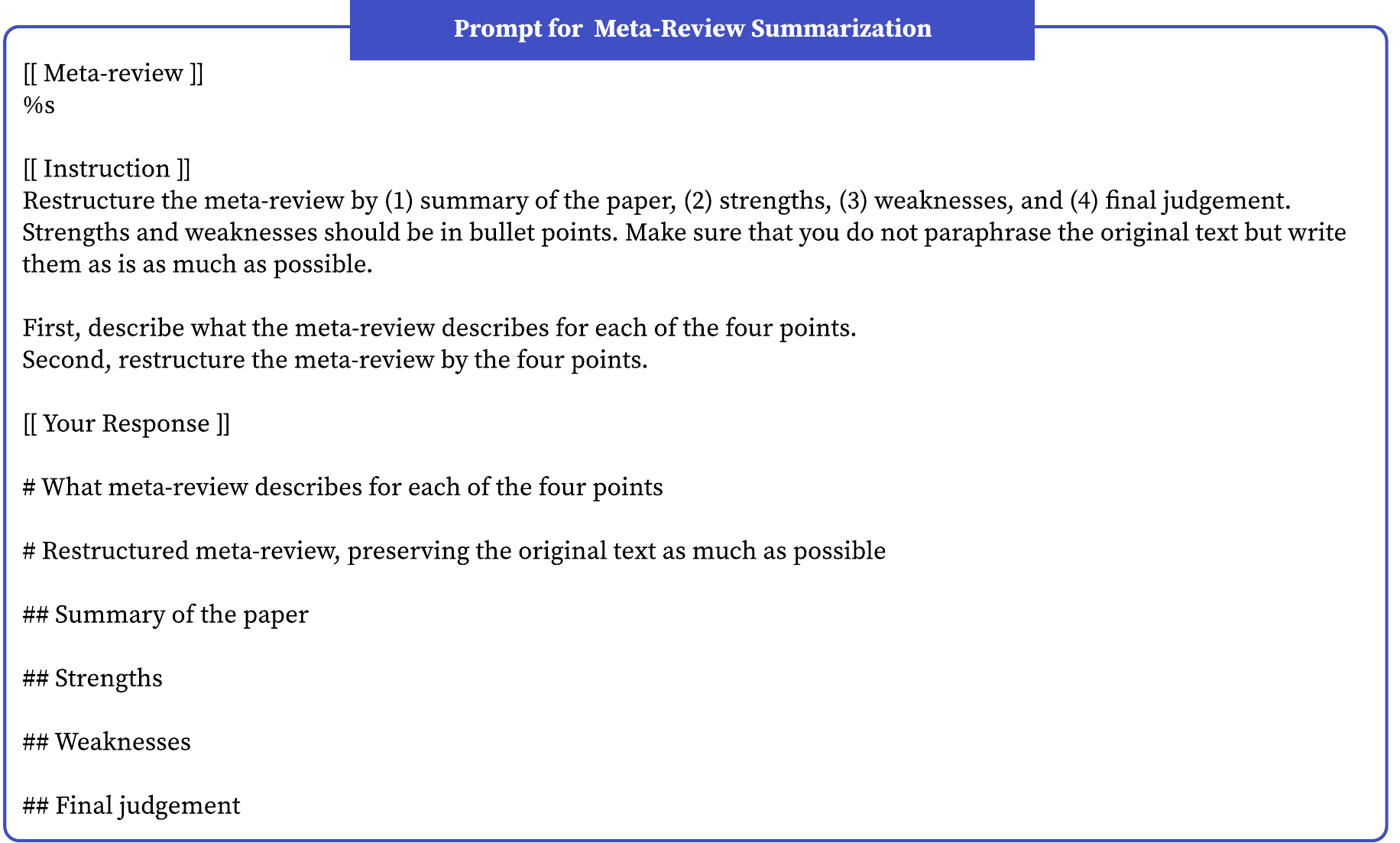}
    \caption{Prompt for Meta-Review Summarization}
    \label{fig:metareview-summarization}
\end{figure*}

\begin{figure*}[h]
    \centering
    \includegraphics[width=\linewidth]{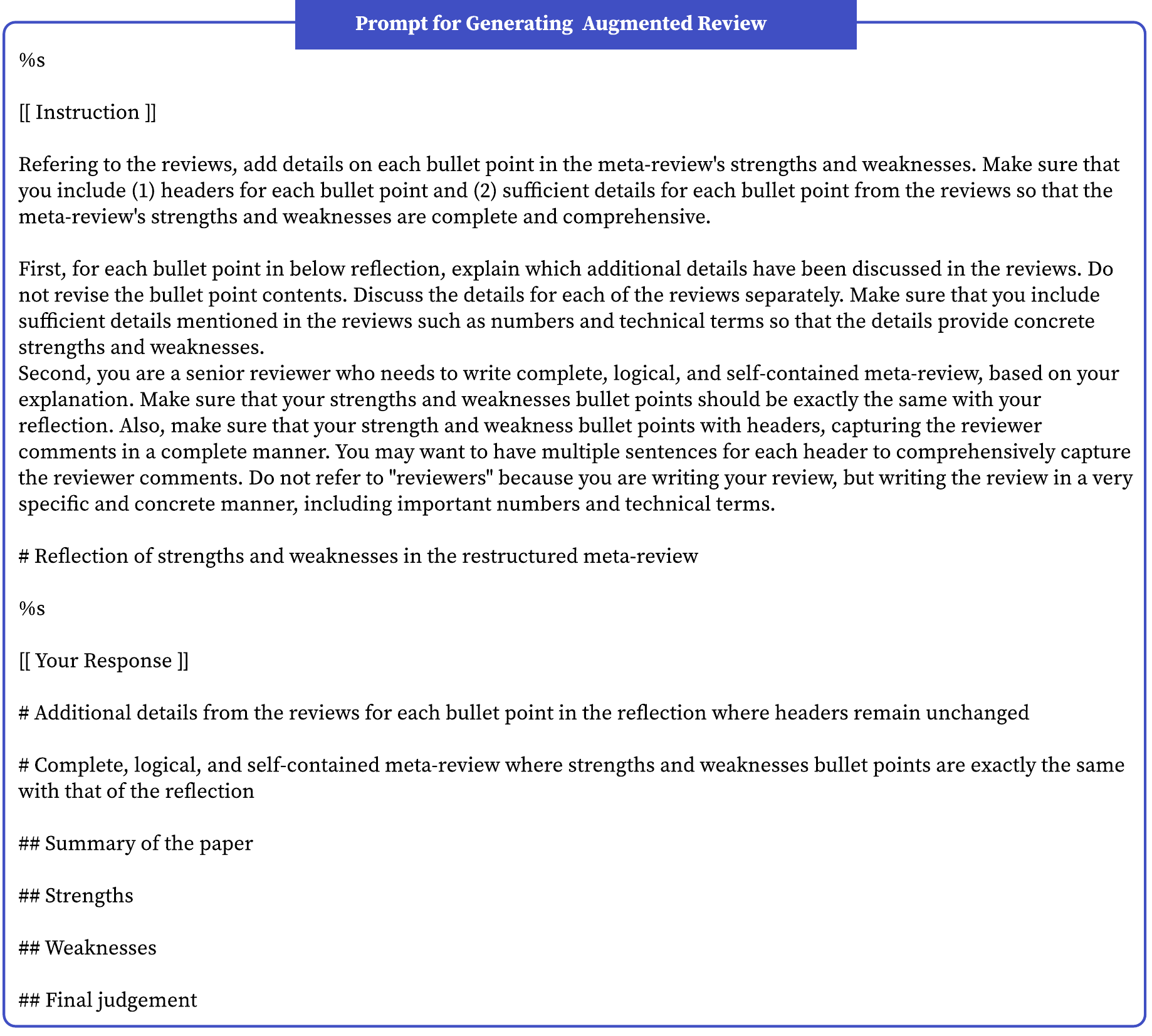}
    \caption{Prompt for Generating  Augmented Review }
    \label{fig:augmentedReview}
\end{figure*}

\begin{figure*}[h]
    \centering
    \includegraphics[width=\linewidth]{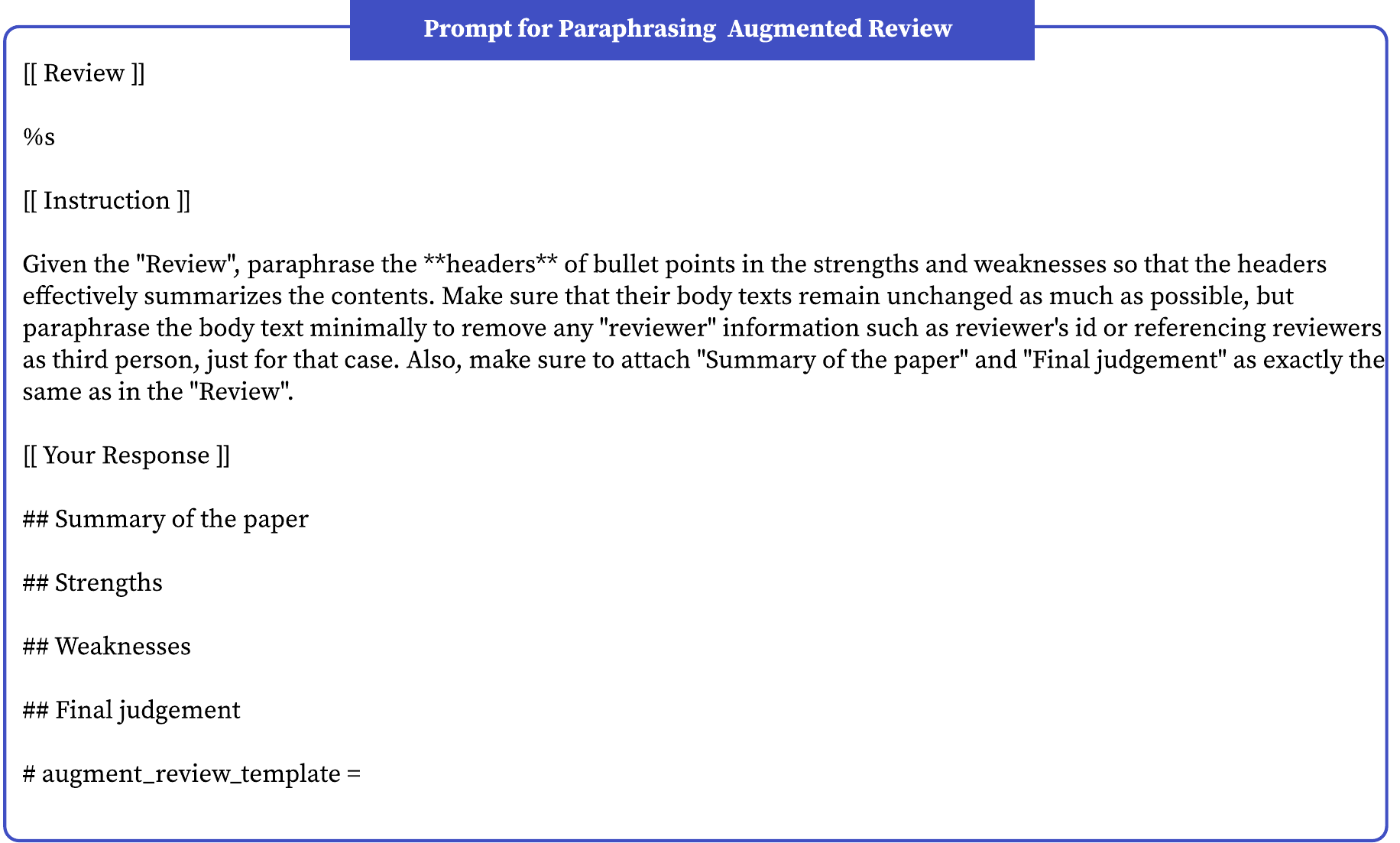}
    \caption{Prompt for Paraphrasing  Augmented Review }
    \label{fig:paraphrasing}
\end{figure*}

\clearpage

% \subsection{Additional Experiments Results}

% In this section, we provide detailed results on the focus of LLM and human reviewers when they are reviewing papers.

% confusion matrix  ggests that the er- rors tend to occur in semantically related categories, indicating that the misclassifications are not arbitrary but rather reflect subtle ambiguities inherent in the data. 

\

\subsubsection{Prompts for LLM Review Generation}

\label{appendix:llm-prompt}
Figure~\ref{fig:llm-review-gen} shows the prompt for using LLM to generate reviews from paper.

\begin{figure*}[h]
    \centering
    \includegraphics[width=1\linewidth]{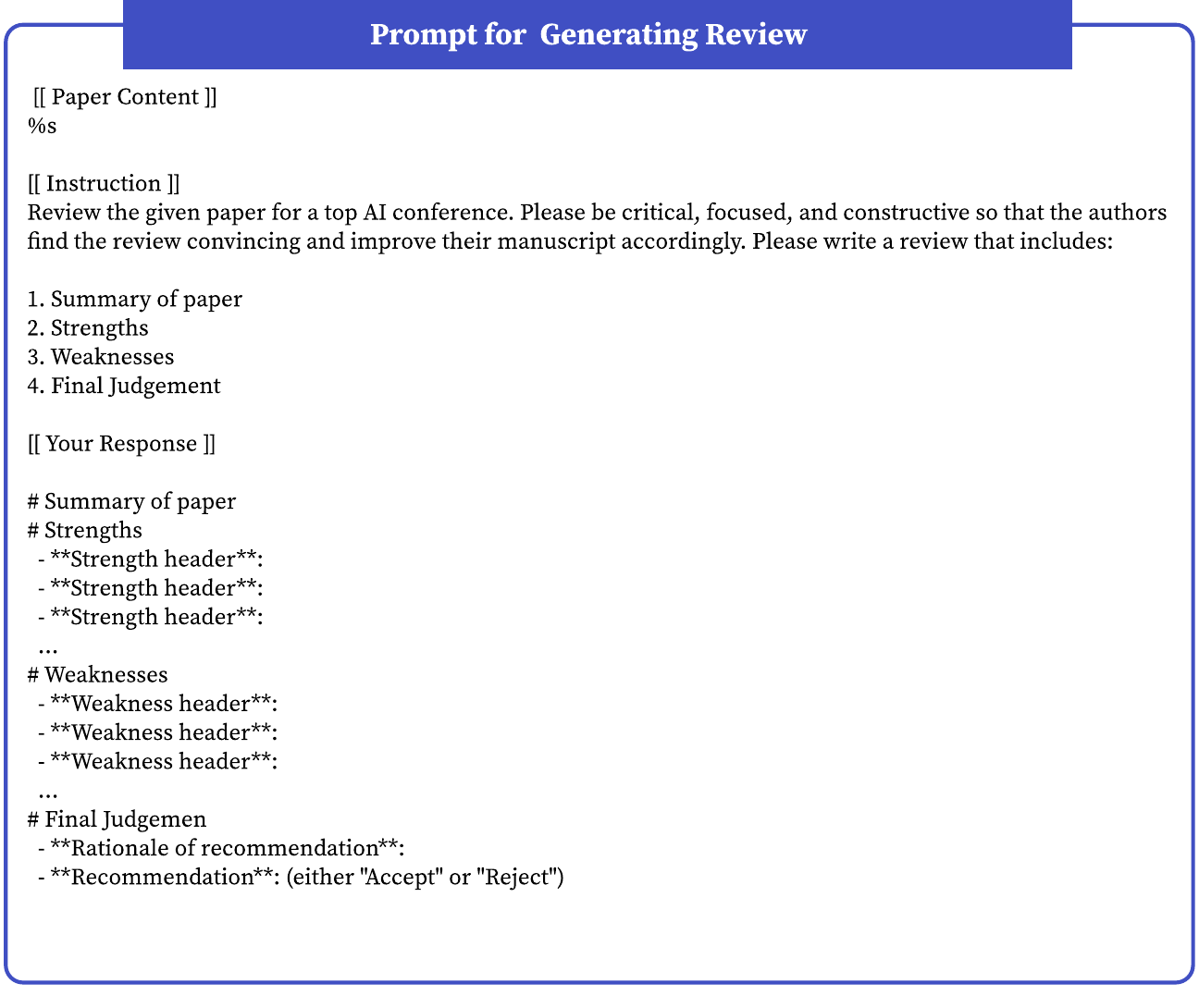}
    \caption{Prompt for LLM Review Generation}
    \label{fig:llm-review-gen}
\end{figure*}

\clearpage

\subsection{Details of Building Automatic Annotator }
\subsubsection{AI paper writing guidelines}
\label{appendix:guidelines}
To ensure guidelines are comprehensive, we collected guidelines from 9 sources, comprising a total of 243 items, as shown in Table~\ref{tab:guideline_item_count}. An item refers to a specific requirement mentioned in the guidelines, which serves as a distinct criterion for reviewing or writing a paper.
\begin{table}[h]
\centering
\caption{Guidelines and Item Count Summary}
\label{tab:guideline_item_count}
\begin{tabular}{lr}
\toprule
Guideline & Item Count \\
\midrule
ICML Paper Writing Best Practices\footnotemark[1] & 38 \\
ICML 2023 Paper Guidelines\footnotemark[2] & 30 \\
NIPS 2024 Reviewer Guidelines\footnotemark[3] & 18 \\
ACL Checklist\footnotemark[4] & 49 \\
How to Write a Good Research Paper in the Machine Learning Area\footnotemark[5] & 6 \\
ACL Ethics Review Questions\footnotemark[6] & 21 \\
AAAI Reproducibility Checklist\footnotemark[7] & 29 \\
NeurIPS 2021 Paper Checklist Guidelines\footnotemark[8] & 46 \\
ICLR 2019 Guidelines\footnotemark[9] & 6 \\
\midrule
\textbf{Total Count} & \textbf{243} \\
\bottomrule
\end{tabular}
\end{table}

\footnotetext[1]{\url{https://icml.cc/Conferences/2022/BestPractices}}
\footnotetext[2]{\url{https://icml.cc/Conferences/2023/PaperGuidelines}}
\footnotetext[3]{\url{https://neurips.cc/Conferences/2024/ReviewerGuidelines}}
\footnotetext[4]{\url{https://aclrollingreview.org/responsibleNLPresearch/}}
\footnotetext[5]{\url{https://www.turing.com/kb/how-to-write-research-paper-in-machine-learning-area}}
\footnotetext[6]{\url{https://2023.eacl.org/ethics/review-questions/}}
\footnotetext[7]{\url{https://aaai.org/conference/aaai/aaai-25/aaai-25-reproducibility-checklist/}}
\footnotetext[8]{\url{https://neurips.cc/Conferences/2021/PaperInformation/PaperChecklist}}
\footnotetext[9]{\url{https://iclr.cc/Conferences/2019/Reviewer_Guidelines}}

\clearpage

\subsubsection{Target and aspect facets}
\label{appendix:target_and_aspect}

\begin{table}[H]
    \centering
    \renewcommand{\arraystretch}{1.2}
    \captionsetup{justification=raggedright, singlelinecheck=false}
    \caption{We aim to analyze focus distributions of LLM reviews based on the targets and aspects. To identify the specific facets for targets (i.e., what the review praises or critiques) and aspects (i.e., the specific elements of the target being evaluated), we surveyed 9 AI paper submission guidelines (Appendix ~\ref{appendix:guidelines}) and prior research on review analysis~\citep{chakraborty2020aspect, ghosal2022peer, yuan2022can}. The facets were used as the codebook for human annotations.}
    \resizebox{\textwidth}{!}{%
    \begin{tabular}{lp{14cm}}
        \hline
        \multicolumn{2}{c}{\textbf{Target}} \\
        \hline
        \textbf{Facet} & \textbf{Definition (The review addresses ...)} \\
        \hline
        \textbf{Problem} & Motivation, task definitions, and problem statements. \\
        \textbf{Prior Research} & References and contextual positioning of the submission. \\
        \textbf{Method} & Proposed approach, techniques, algorithms, or datasets. \\
        \textbf{Theory} & Theoretical foundations, assumptions, proofs, or justifications. \\
        \textbf{Experiment} & Experimental setup, results, and analysis. \\
        \textbf{Conclusion} & Findings, implications, discussions, and takeaways. \\
        \textbf{Paper} & General targets of the paper without specifying a particular target \\
        \hline
        \multicolumn{2}{c}{\textbf{Aspect}} \\
        \hline
        \textbf{Facet} & \textbf{Definition (The review addresses ...)} \\
        \hline
        \textbf{Impact} & Significance or practical influence of the work. \\
        \textbf{Novelty} & Originality of the submission compared to prior research. \\
        \textbf{Clarity} & Readability, ambiguity, or communication aspects. \\
        \textbf{Validity} & Soundness, completeness, and rigor. \\
        \textbf{Not-specific} & Multiple targets without emphasis on a particular aspect. \\
        \hline
    \end{tabular}%
    }
    
\end{table}

\subsubsection{Prompts}
\label{appendix:annotate-prompt}

In this section, we provide prompts designed to annotate reviews. We designed 4 prompts where each corresponds to one of the four combinations of target/aspect and strength/weakness. Specifically, we designed Target-Strength (Figure~\ref{fig:target-strength-prompt}), Aspect-Strength, (Figure~\ref{fig:aspect-strength-prompt}), Target-Weakness (Figure~\ref{fig:target-weakness-prompt}) , and Aspect-Weakness (Figure~\ref{fig:aspect-weakness-prompt}) prompts.

% Strengths and weaknesses identification prompt 

\begin{figure*}[t]
    \centering
    \includegraphics[width=\linewidth, height=1\textheight, keepaspectratio]{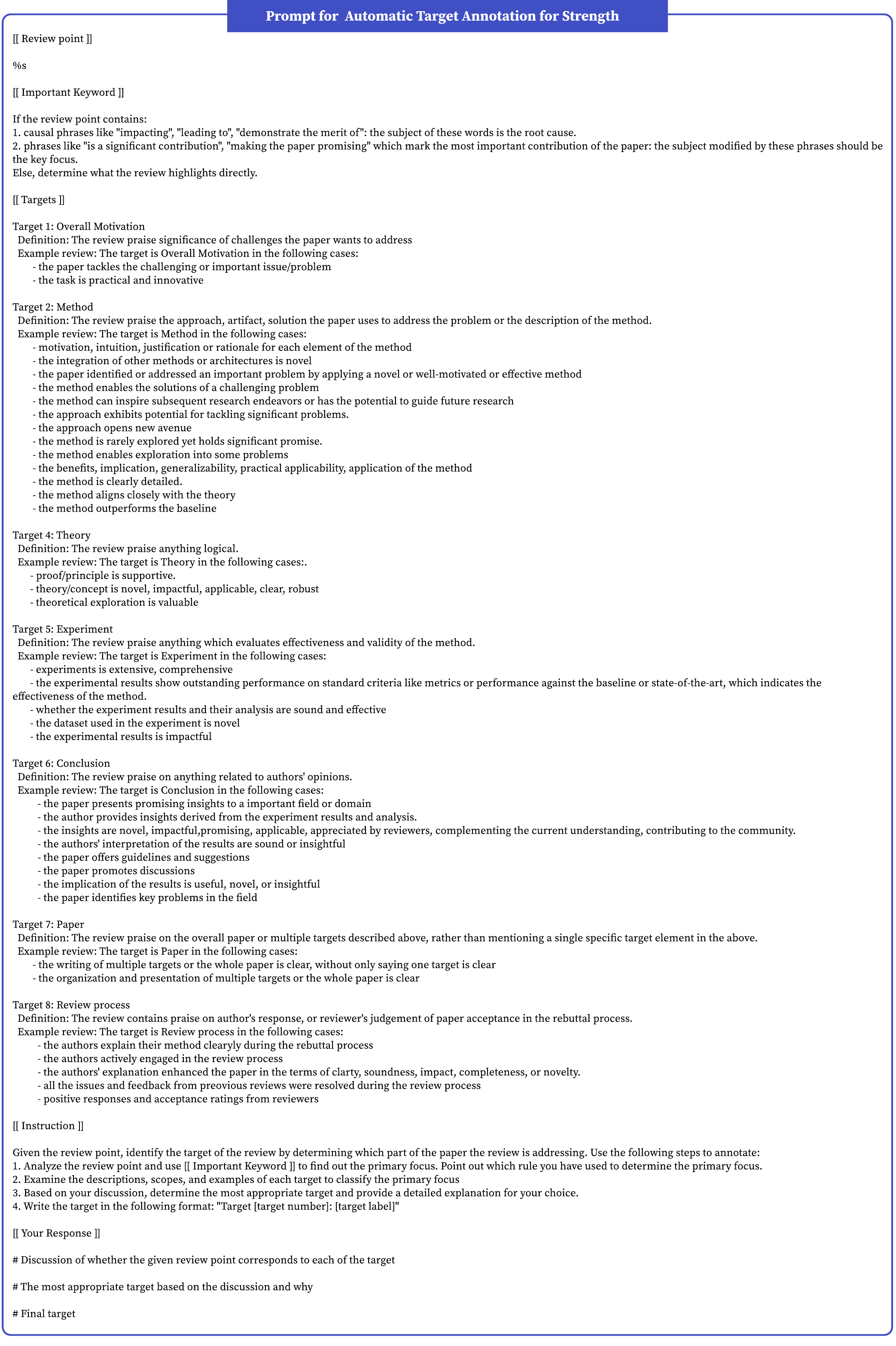}
    \caption{Prompt for Automatic Target Annotation for Strength}
    \label{fig:target-strength-prompt}
\end{figure*}

\begin{figure*}[t]
    \centering
    \includegraphics[width=\linewidth, height=1\textheight, keepaspectratio]{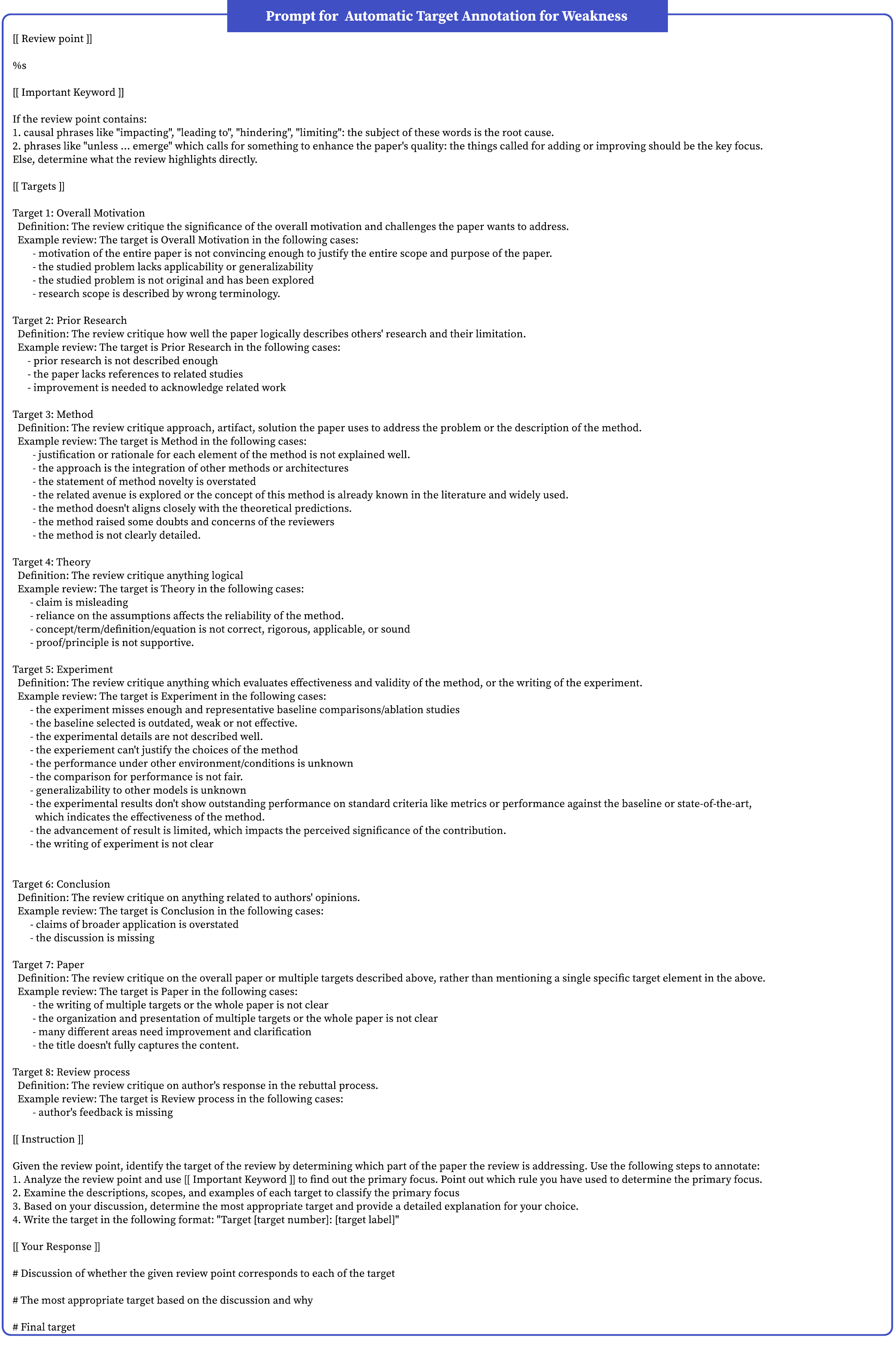}
    \caption{Prompt for Automatic Target Annotation for Weakness}
    \label{fig:target-weakness-prompt}
\end{figure*}

\begin{figure*}[t]
    \centering
    \includegraphics[width=\linewidth, height=1\textheight, keepaspectratio]{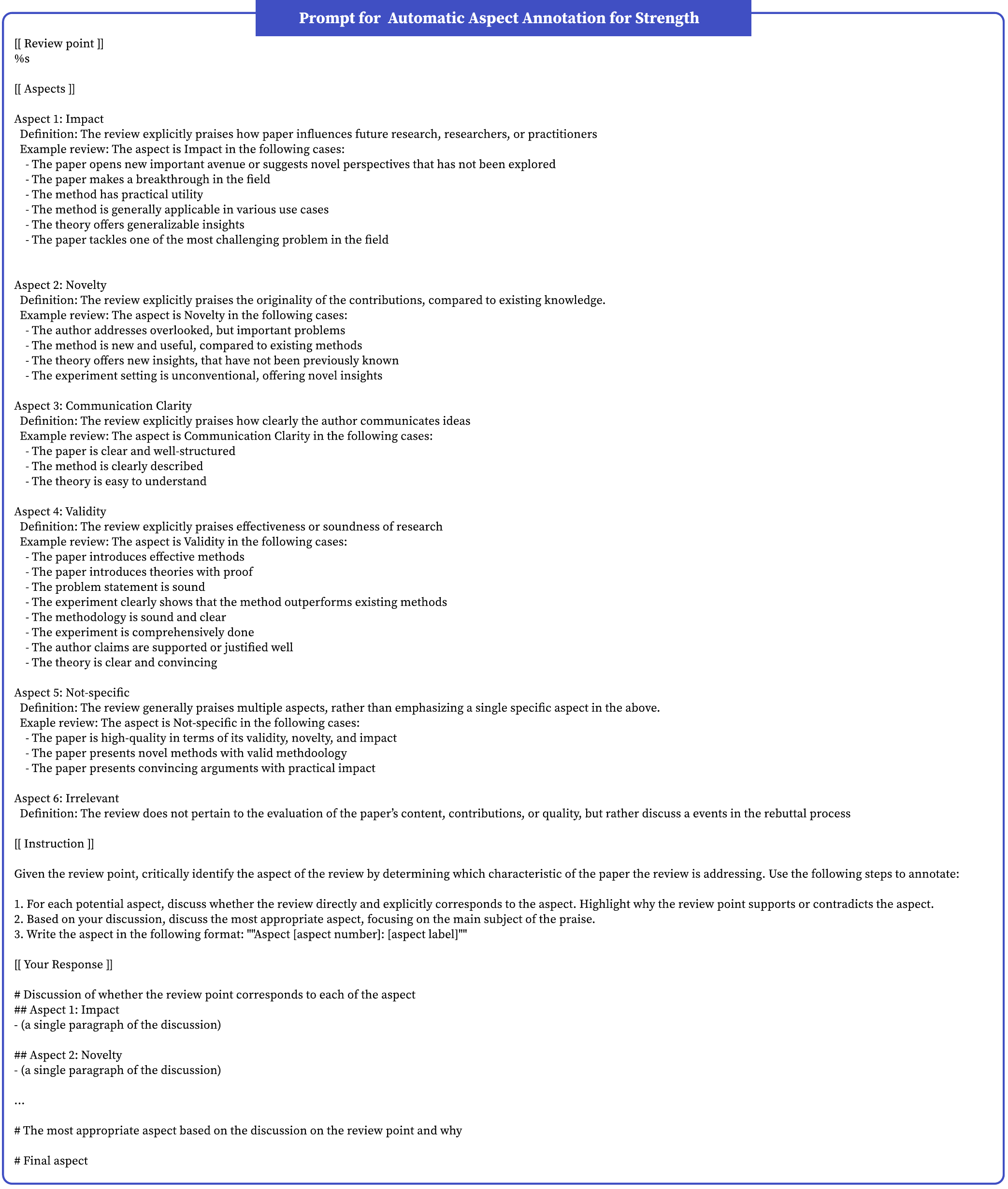}
    \caption{Prompt for Automatic Aspect Annotation for Strength}
    \label{fig:aspect-strength-prompt}
\end{figure*}

\begin{figure*}[t]
    \centering
    \includegraphics[width=\linewidth, height=1\textheight, keepaspectratio]{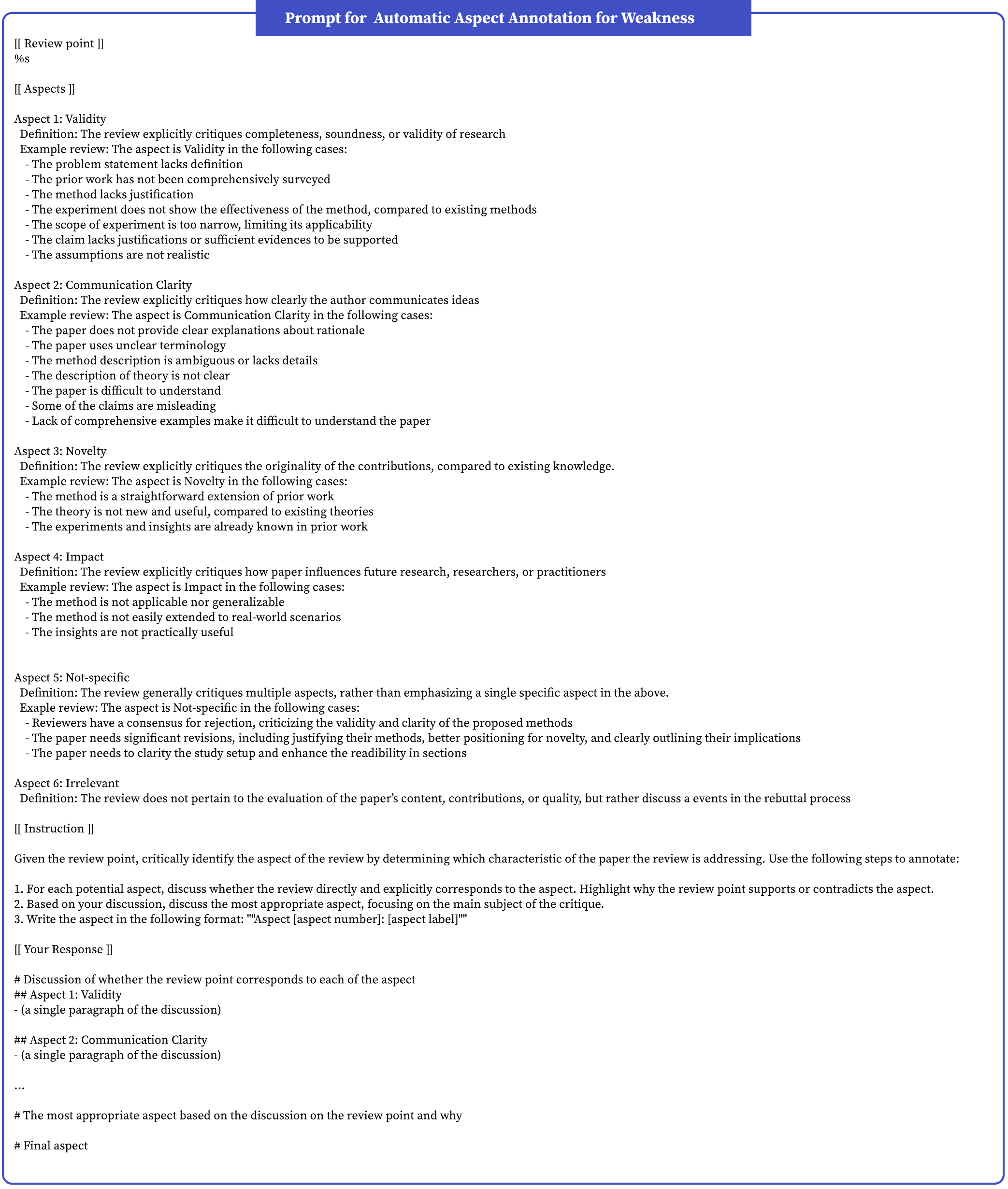}
    \caption{Prompt for Automatic Aspect Annotation for Weakness}
    \label{fig:aspect-weakness-prompt}
\end{figure*}

\clearpage

\subsubsection{Annotation Comparison}
\label{appendix:annotate-confusion-matrix}

We present a comparison between LLM and human annotations for both target and aspect. Figures~\ref{fig:error-target-matrix} and Figure~\ref{fig:error-aspect-matrix} illustrate the discrepancies. Areas of alignment between LLM and human annotations are shown in green, while red highlights regions with significant discrepancies.

\begin{figure}[h]
    \centering
    \includegraphics[width=0.5\linewidth]{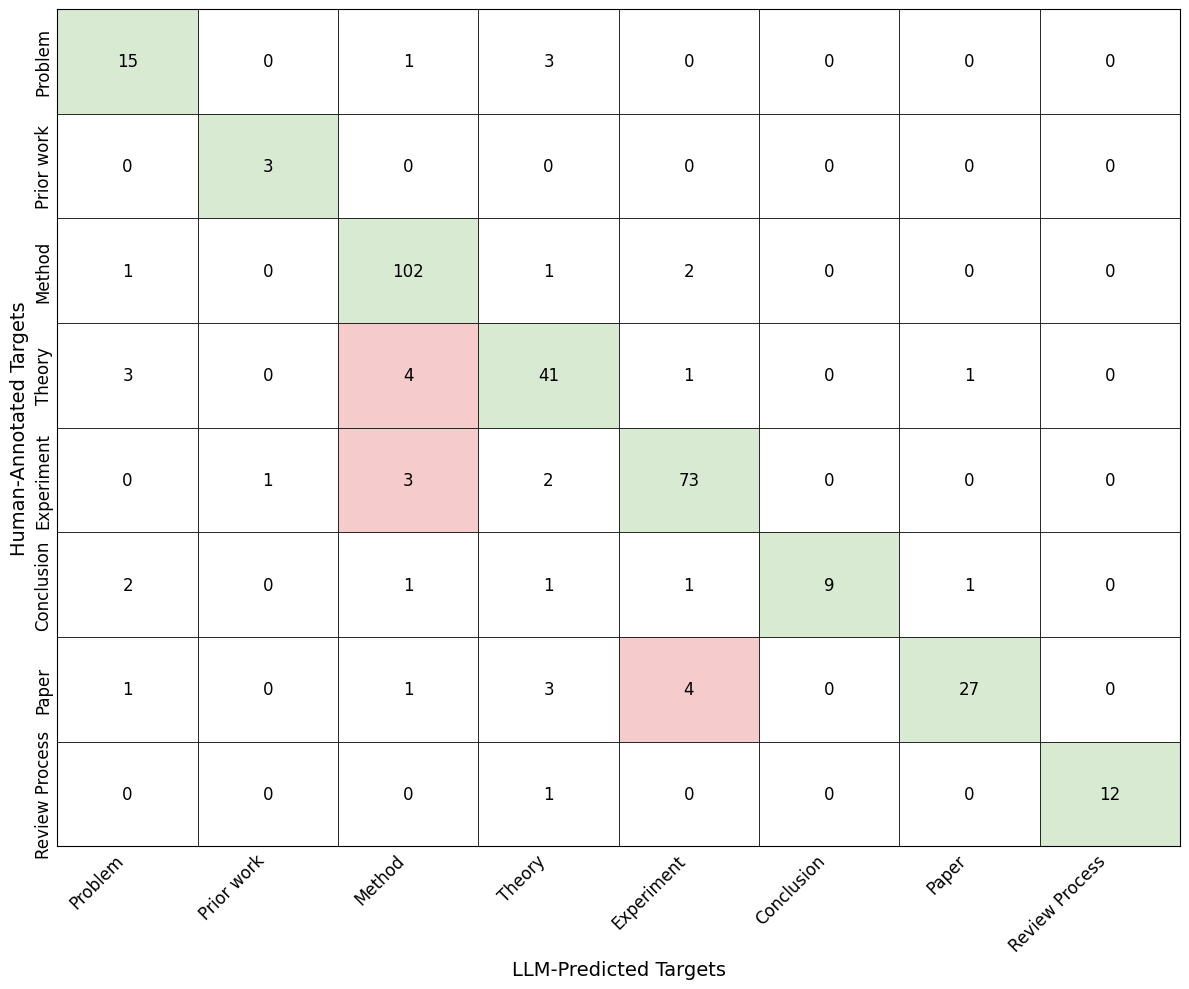}
    \caption{LLM vs. human target annotation}
    \label{fig:error-target-matrix}
\end{figure}

\begin{figure}[h]
    \centering
    \includegraphics[width=0.5\linewidth]{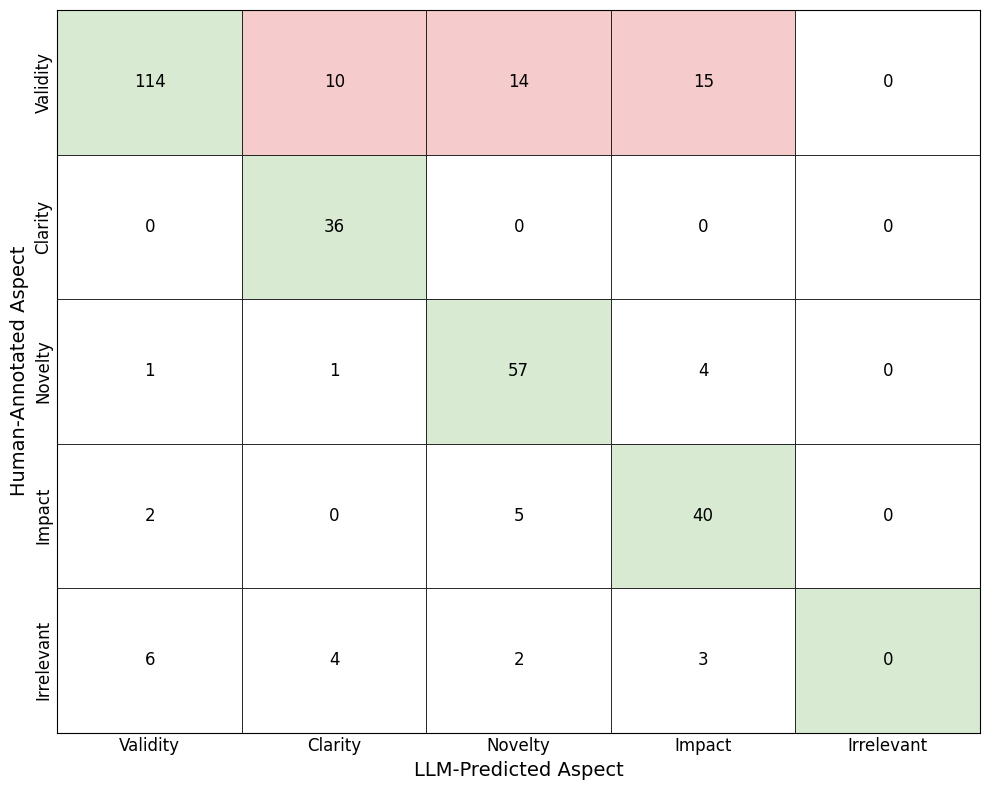}
    \caption{LLM vs. human aspect annotation}
    \label{fig:error-aspect-matrix}
\end{figure}

While LLM annotations differ from human annotations in some cases, certain discrepancies remain reasonable. Figure~\ref{fig:target-case} and Figure~\ref{fig:aspect-case}  illustrate examples of such reasonable discrepancies.

\begin{figure}[h]
    \centering
    \includegraphics[width=1\linewidth]{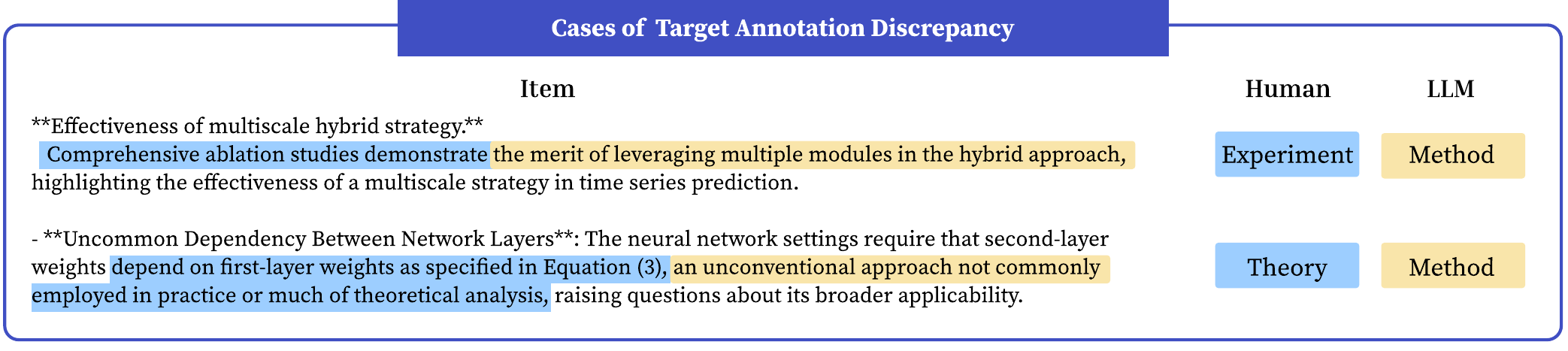}
    \caption{Cases of Target Annotation Discrepancy}
    \label{fig:target-case}

\end{figure}

\begin{figure}[h]
    \centering
    \includegraphics[width=1\linewidth]{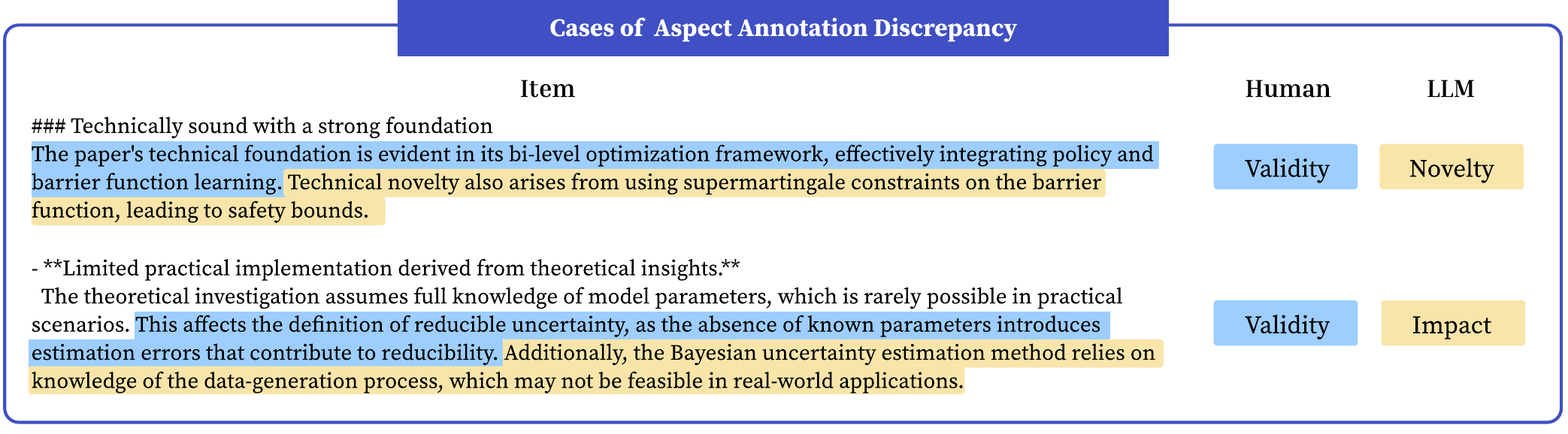}
    \caption{Cases of  Aspect Annotation Discrepancy}
    \label{fig:aspect-case}

\end{figure}

\clearpage

\subsection{Fine-Tuning Details}
\label{appendix:finetune}
\subsubsection{Fine-Tuning Dataset Construction}
We constructed the fine-tuning dataset based on the corpus of  papers described in Section~\ref{sec: pipeline}. We retained 582 training samples and 98 test samples. 5 samples were excluded during tokenization due to exceeding the model's maximum token length
\subsubsection{Fine-Tuning Method}
We employed supervised fine-tuning (SFT) to adapt the GPT-4o base model to our task-specific objectives. Fine-tuning was conducted using the OpenAI Fine-Tuning API\footnote{\url{https://platform.openai.com/docs/api-reference/fine-tuning}}, which abstracts away hardware and infrastructure details. Therefore, we do not report GPU type or compute hours. Table~\ref{tab: fine-tuning config} summarizes the hyperparameter configuration used during training.

\begin{table}[htbp]
\centering
\caption{Hyperparameter settings for supervised fine-tuning.}
\label{tab: fine-tuning config}
\begin{tabular}{lc}
\toprule[1.5pt]
\textbf{Parameter} & \textbf{Value} \\
\midrule
total epochs & 4 \\
batch size & 4 \\
learning rate multiplier & 0.1 \\
\bottomrule[1.5pt]
\end{tabular}
\end{table}
\subsection{Detailed Evaluation Results}
\label{appendix:result}

The following tables present a comprehensive performance comparison of models across different metrics and evaluation targets, including both strengths and weaknesses (Table~\ref{tab:target-all}), as well as separate analyses focusing on strengths (Table~\ref{tab:target_strength_metrics}) and weaknesses (Table~\ref{tab:target_weakness_metrics}). Additionally, we provide a similar comparison across metrics and broader aspects, including both strengths and weaknesses (Table~\ref{tab:aspect-all}), strengths alone (Table~\ref{fig:aspect-strength}), and weaknesses alone (Table~\ref{fig:aspect-weakness}).

\begin{table}[h]
    \centering
    \caption{Performance Comparison of Models Across Metrics and Targets (Including both Strengths and Weaknesses)}    
    \small
    \begin{tabular}{lccccccc}
        \toprule
        Target & Problem & Prior Research & Method & Theory & Experiment & Conclusion & Paper \\
        \midrule
        F1 (gpt-4o-mini) & 0.268 & 0.076 & 0.737 & 0.427 & 0.680 & 0.103 & 0.227 \\
        F1 (gpt-4o) & 0.292 & 0.052 & 0.741 & 0.448 & 0.673 & 0.089 & 0.247 \\
        F1 (o1-mini) & 0.275 & 0.054 & \textbf{0.764} & 0.472 & \textbf{0.684} & \textbf{0.175} & \textbf{0.253} \\
        F1 (o1) & 0.274 & 0.044 & 0.754 & \textbf{0.489} & 0.673 & 0.133 & 0.091 \\
        F1 (llama-70B) & 0.269 & 0.049 & 0.711 & 0.410 & 0.659 & 0.172 & 0.158 \\
        F1 (llama-405B) & 0.158 & 0.031 & 0.690 & 0.427 & 0.662 & 0.167 & 0.134 \\
        F1 (deepseek-r1) & \textbf{0.297} & \textbf{0.081} & 0.729 & 0.473 & 0.682 & 0.164 & 0.152 \\
        F1 (deepseek-v3) & 0.241 & 0.051 & 0.725 & 0.405 & 0.680 & 0.110 & 0.092 \\
        \midrule
        Prec (gpt-4o-mini) & 0.317 & 0.134 & 0.647 & 0.317 & 0.549 & 0.063 & 0.241 \\
        Prec (gpt-4o) & 0.298 & 0.109 & 0.634 & 0.334 & 0.547 & 0.057 & 0.251 \\
        Prec (o1-mini) & 0.315 & 0.130 & 0.639 & 0.342 & 0.549 & 0.107 & 0.274 \\
        Prec (o1) & 0.279 & 0.064 & \textbf{0.648} & \textbf{0.381} & 0.549 & 0.111 & 0.245 \\
        Prec (llama-70B) & \textbf{0.339} & \textbf{0.143} & 0.653 & 0.295 & 0.548 & 0.105 & 0.289 \\
        Prec (llama-405B) & 0.324 & 0.071 & 0.647 & 0.310 & \textbf{0.558} & 0.115 & 0.233 \\
        Prec (deepseek-r1) & 0.321 & 0.099 & 0.639 & 0.327 & 0.549 & \textbf{0.135} & \textbf{0.301} \\
        Prec (deepseek-v3) & 0.288 & 0.100 & 0.645 & 0.280 & 0.547 & 0.076 & 0.249 \\
        \midrule
        Rec (gpt-4o-mini) & 0.233 & 0.053 & 0.870 & 0.691 & 0.983 & 0.274 & 0.232 \\
        Rec (gpt-4o) & 0.297 & 0.034 & 0.899 & 0.723 & 0.965 & 0.202 & \textbf{0.270} \\
        Rec (o1-mini) & 0.266 & 0.034 & \textbf{0.952} & 0.834 & \textbf{0.994} & \textbf{0.536} & 0.249 \\
        Rec (o1) & \textbf{0.353} & 0.034 & 0.905 & 0.736 & 0.963 & 0.167 & 0.056 \\
        Rec (llama-70B) & 0.246 & 0.030 & 0.803 & 0.720 & 0.919 & 0.476 & 0.146 \\
        Rec (llama-405B) & 0.108 & 0.020 & 0.774 & 0.694 & 0.894 & 0.300 & 0.095 \\
        Rec (deepseek-r1) & 0.299 & \textbf{0.069} & 0.859 & \textbf{0.865} & 0.983 & 0.357 & 0.102 \\
        Rec (deepseek-v3) & 0.210 & 0.035 & 0.844 & 0.755 & 0.981 & 0.238 & 0.058 \\
        \bottomrule
    \end{tabular}

    \label{tab:target-all}
\end{table}

\begin{table}[h]
    \centering
    \small
    \caption{Performance Comparison of Models Across Metrics and Targets (Strengths)}
    
    \begin{tabular}{lccccccc}
        \toprule
        Target & Problem & Prior Research & Method & Theory & Experiment & Conclusion & Paper \\
        \midrule
        F1 (gpt-4o-mini) & 0.283 & 0.000 & \textbf{0.760} & 0.424 & 0.511 & 0.118 & 0.232 \\
        F1 (gpt-4o) & 0.329 & 0.000 & 0.756 & 0.446 & \textbf{0.517} & 0.143 & 0.119 \\
        F1 (o1-mini) & 0.345 & 0.000 & 0.753 & 0.411 & 0.511 & 0.300 & \textbf{0.233} \\
        F1 (o1) & 0.384 & 0.000 & 0.749 & \textbf{0.470} & 0.512 & 0.267 & 0.061 \\
        F1 (llama-70B) & 0.245 & 0.000 & 0.750 & 0.420 & 0.516 & 0.242 & 0.198 \\
        F1 (llama-405B) & 0.160 & 0.000 & 0.755 & 0.455 & 0.516 & \textbf{0.333} & 0.079 \\
        F1 (deepseek-r1) & \textbf{0.396} & 0.000 & 0.749 & 0.436 & 0.513 & 0.174 & 0.135 \\
        F1 (deepseek-v3) & 0.331 & 0.000 & 0.755 & 0.423 & 0.509 & 0.114 & 0.086 \\
        \midrule
        Prec (gpt-4o-mini) & 0.315 & 0.000 & 0.622 & 0.286 & 0.343 & 0.071 & 0.198 \\
        Prec (gpt-4o) & 0.295 & 0.000 & 0.616 & 0.299 & 0.350 & 0.091 & 0.182 \\
        Prec (o1-mini) & 0.314 & 0.000 & 0.611 & 0.264 & 0.343 & 0.176 & 0.203 \\
        Prec (o1) & 0.285 & 0.000 & \textbf{0.624} & \textbf{0.322} & 0.346 & 0.222 & 0.172 \\
        Prec (llama-70B) & 0.404 & 0.000 & 0.620 & 0.275 & 0.352 & 0.148 & 0.178 \\
        Prec (llama-405B) & \textbf{0.419} & 0.000 & 0.620 & 0.319 & \textbf{0.358} & \textbf{0.231} & 0.163 \\
        Prec (deepseek-r1) & 0.355 & 0.000 & 0.617 & 0.289 & 0.347 & 0.103 & \textbf{0.279} \\
        Prec (deepseek-v3) & 0.364 & 0.000 & 0.620 & 0.276 & 0.344 & 0.069 & 0.154 \\
        \midrule
        Rec (gpt-4o-mini) & 0.258 & 0.000 & 0.975 & 0.819 & \textbf{0.996} & 0.333 & \textbf{0.281} \\
        Rec (gpt-4o) & 0.371 & 0.000 & 0.978 & 0.872 & 0.991 & 0.333 & 0.089 \\
        Rec (o1-mini) & 0.382 & 0.000 & \textbf{0.980} & \textbf{0.935} & \textbf{0.996} & \textbf{1.000} & 0.274 \\
        Rec (o1) & \textbf{0.588} & 0.000 & 0.936 & 0.872 & 0.987 & 0.333 & 0.037 \\
        Rec (llama-70B) & 0.176 & 0.000 & 0.948 & 0.894 & 0.969 & 0.667 & 0.224 \\
        Rec (llama-405B) & 0.099 & 0.000 & 0.965 & 0.796 & 0.921 & 0.600 & 0.052 \\
        Rec (deepseek-r1) & 0.447 & 0.000 & 0.953 & 0.883 & 0.983 & 0.571 & 0.089 \\
        Rec (deepseek-v3) & 0.303 & 0.000 & 0.963 & 0.904 & 0.982 & 0.333 & 0.059 \\
        \bottomrule
    \end{tabular}
    \label{tab:target_strength_metrics}
\end{table}

\begin{table}[h]
    \centering
    \small
    \caption{Performance Comparison of Models Across Metrics and Targets (Weaknesses)}
    
    \begin{tabular}{lccccccc}
        \toprule
        Target & Problem & Prior Research & Method & Theory & Experiment & Conclusion & Paper \\
        \midrule
        F1 (gpt-4o-mini) & 0.253 & 0.153 & 0.715 & 0.430 & 0.849 & 0.088 & 0.222 \\
        F1 (gpt-4o) & 0.256 & 0.104 & 0.726 & 0.449 & 0.830 & 0.036 & \textbf{0.375} \\
        F1 (o1-mini) & 0.204 & 0.108 & \textbf{0.774} & 0.534 & \textbf{0.857} & 0.050 & 0.272 \\
        F1 (o1) & 0.164 & 0.089 & 0.760 & 0.508 & 0.835 & 0.000 & 0.120 \\
        F1 (llama-70B) & \textbf{0.294} & 0.098 & 0.672 & 0.400 & 0.802 & 0.103 & 0.118 \\
        F1 (llama-405B) & 0.155 & 0.062 & 0.625 & 0.399 & 0.809 & 0.000 & 0.190 \\
        F1 (deepseek-r1) & 0.198 & \textbf{0.163} & 0.709 & \textbf{0.510} & 0.852 & \textbf{0.154} & 0.169 \\
        F1 (deepseek-v3) & 0.151 & 0.103 & 0.696 & 0.387 & 0.850 & 0.105 & 0.099 \\
        \midrule
        Prec (gpt-4o-mini) & \textbf{0.320} & 0.268 & 0.672 & 0.347 & 0.755 & 0.056 & 0.283 \\
        Prec (gpt-4o) & 0.301 & 0.219 & 0.651 & 0.369 & 0.743 & 0.024 & 0.321 \\
        Prec (o1-mini) & 0.315 & 0.259 & 0.666 & 0.420 & 0.754 & 0.038 & 0.345 \\
        Prec (o1) & 0.273 & 0.127 & 0.672 & \textbf{0.440} & 0.752 & 0.000 & 0.317 \\
        Prec (llama-70B) & 0.274 & \textbf{0.286} & \textbf{0.687} & 0.315 & 0.744 & 0.062 & \textbf{0.400} \\
        Prec (llama-405B) & 0.228 & 0.143 & 0.673 & 0.300 & \textbf{0.758} & 0.000 & 0.304 \\
        Prec (deepseek-r1) & 0.287 & 0.197 & 0.661 & 0.365 & 0.750 & \textbf{0.167} & 0.323 \\
        Prec (deepseek-v3) & 0.212 & 0.200 & 0.669 & 0.284 & 0.750 & 0.083 & 0.345 \\
        \midrule
        Rec (gpt-4o-mini) & 0.209 & 0.107 & 0.764 & 0.563 & 0.970 & \textbf{0.214} & 0.183 \\
        Rec (gpt-4o) & 0.222 & 0.068 & 0.821 & 0.574 & 0.939 & 0.071 & \textbf{0.451} \\
        Rec (o1-mini) & 0.151 & 0.068 & \textbf{0.924} & 0.732 & \textbf{0.992} & 0.071 & 0.224 \\
        Rec (o1) & 0.118 & 0.068 & 0.874 & 0.600 & 0.939 & 0.000 & 0.074 \\
        Rec (llama-70B) & \textbf{0.316} & 0.059 & 0.658 & 0.547 & 0.869 & 0.286 & 0.069 \\
        Rec (llama-405B) & 0.118 & 0.040 & 0.583 & 0.593 & 0.867 & 0.000 & 0.138 \\
        Rec (deepseek-r1) & 0.151 & \textbf{0.139} & 0.764 & \textbf{0.847} & 0.984 & 0.143 & 0.115 \\
        Rec (deepseek-v3) & 0.118 & 0.069 & 0.725 & 0.605 & 0.980 & 0.143 & 0.057 \\
        \bottomrule
    \end{tabular}
    \label{tab:target_weakness_metrics}
\end{table}

\begin{table}[h]
    \centering
    \caption{Performance Comparison of Models Across Metrics and Aspects (Including both Strengths and Weaknesses)}
    \small
    \begin{tabular}{lcccc}
        \toprule
        Aspect & Novelty & Impact & Validity & Clarity \\
        \midrule
        F1 (gpt-4o-mini)   & 0.334 & 0.390 & \textbf{0.775} & 0.396 \\
        F1 (gpt-4o)        & 0.378 & \textbf{0.428} & 0.769 & 0.365 \\
        F1 (o1-mini)       & 0.386 & 0.427 & 0.773 & 0.395 \\
        F1 (o1)            & \textbf{0.404} & 0.399 & 0.772 & \textbf{0.401} \\
        F1 (llama-70B)     & 0.334 & 0.322 & 0.769 & 0.327 \\
        F1 (llama-405B)    & 0.337 & 0.318 & 0.772 & 0.278 \\
        F1 (deepseek-r1)   & 0.387 & 0.414 & \textbf{0.775} & 0.266 \\
        F1 (deepseek-v3)   & 0.346 & 0.422 & 0.768 & 0.187 \\
        \midrule
        Prec (gpt-4o-mini) & 0.367 & 0.291 & \textbf{0.671} & 0.317 \\
        Prec (gpt-4o)      & 0.474 & 0.313 & 0.668 & 0.298 \\
        Prec (o1-mini)     & 0.528 & 0.300 & 0.668 & 0.311 \\
        Prec (o1)          & 0.589 & 0.305 & 0.669 & 0.334 \\
        Prec (llama-70B)   & \textbf{0.665} & \textbf{0.318} & 0.667 & 0.337 \\
        Prec (llama-405B)  & 0.587 & 0.302 & \textbf{0.671} & 0.332 \\
        Prec (deepseek-r1) & 0.535 & 0.308 & 0.670 & \textbf{0.339} \\
        Prec (deepseek-v3) & 0.504 & 0.306 & 0.664 & 0.309 \\
        \midrule
        Rec (gpt-4o-mini)  & 0.460 & 0.600 & \textbf{0.990} & \textbf{0.549} \\
        Rec (gpt-4o)       & 0.506 & 0.689 & 0.975 & 0.485 \\
        Rec (o1-mini)      & \textbf{0.507} & \textbf{0.758} & \textbf{0.990} & 0.548 \\
        Rec (o1)           & 0.435 & 0.579 & 0.981 & 0.511 \\
        Rec (llama-70B)    & 0.450 & 0.371 & 0.981 & 0.346 \\
        Rec (llama-405B)   & 0.478 & 0.352 & 0.978 & 0.241 \\
        Rec (deepseek-r1)  & 0.502 & 0.632 & 0.988 & 0.219 \\
        Rec (deepseek-v3)  & 0.478 & 0.683 & 0.982 & 0.134 \\
        \bottomrule
    \end{tabular}
    \label{tab:aspect-all}
\end{table}

\begin{table}[h]
    \centering
    \caption{Performance Comparison of Models Across Metrics and Aspects (Strengths)}
    \small
    \begin{tabular}{lcccc}
        \toprule
        Aspect & Novelty & Impact & Validity & Clarity \\
        \midrule
        F1 (gpt-4o-mini)   & 0.643 & 0.474 & \textbf{0.599} & 0.309 \\
        F1 (gpt-4o)        & 0.654 & 0.520 & 0.593 & 0.202 \\
        F1 (o1-mini)       & 0.656 & \textbf{0.556} & 0.592 & 0.299 \\
        F1 (o1)            & 0.626 & 0.530 & 0.596 & \textbf{0.342} \\
        F1 (llama-70B)     & 0.636 & 0.411 & 0.593 & 0.292 \\
        F1 (llama-405B)    & \textbf{0.660} & 0.345 & 0.596 & 0.157 \\
        F1 (deepseek-r1)   & 0.655 & 0.536 & 0.598 & 0.170 \\
        F1 (deepseek-v3)   & \textbf{0.660} & 0.547 & 0.585 & 0.122 \\
        \midrule
        Prec (gpt-4o-mini) & 0.498 & 0.368 & \textbf{0.431} & 0.222 \\
        Prec (gpt-4o)      & 0.498 & 0.398 & 0.428 & 0.190 \\
        Prec (o1-mini)     & 0.501 & 0.403 & 0.424 & 0.224 \\
        Prec (o1)          & \textbf{0.530} & 0.412 & 0.430 & \textbf{0.261} \\
        Prec (llama-70B)   & 0.497 & \textbf{0.467} & 0.426 & 0.236 \\
        Prec (llama-405B)  & 0.506 & 0.368 & \textbf{0.431} & 0.215 \\
        Prec (deepseek-r1) & 0.503 & 0.400 & \textbf{0.431} & 0.224 \\
        Prec (deepseek-v3) & 0.509 & 0.403 & 0.419 & 0.207 \\
        \midrule
        Rec (gpt-4o-mini)  & 0.907 & 0.667 & \textbf{0.986} & \textbf{0.511} \\
        Rec (gpt-4o)       & \textbf{0.955} & 0.749 & 0.965 & 0.216 \\
        Rec (o1-mini)      & 0.949 & \textbf{0.897} & 0.979 & 0.449 \\
        Rec (o1)           & 0.763 & 0.744 & 0.969 & 0.496 \\
        Rec (llama-70B)    & 0.883 & 0.366 & 0.976 & 0.384 \\
        Rec (llama-405B)   & 0.949 & 0.324 & 0.969 & 0.123 \\
        Rec (deepseek-r1)  & 0.937 & 0.809 & 0.976 & 0.137 \\
        Rec (deepseek-v3)  & 0.940 & 0.851 & 0.965 & 0.086 \\
        \bottomrule
    \end{tabular}
    \label{fig:aspect-strength}

\end{table}

\begin{table}[h]
    \centering
    \small
    \caption{Performance Comparison of Models Across Metrics and Aspects (Weaknesses)}
    \begin{tabular}{lcccc}
        \toprule
        Aspect & Novelty & Impact & Validity & Clarity \\
        \midrule
        F1 (gpt-4o-mini)   & 0.024 & 0.306 & 0.951 & 0.484 \\
        F1 (gpt-4o)        & 0.103 & \textbf{0.335} & 0.945 & \textbf{0.528} \\
        F1 (o1-mini)       & 0.116 & 0.299 & \textbf{0.954} & 0.492 \\
        F1 (o1)            & \textbf{0.182} & 0.268 & 0.949 & 0.459 \\
        F1 (llama-70B)     & 0.032 & 0.233 & 0.945 & 0.362 \\
        F1 (llama-405B)    & 0.013 & 0.291 & 0.947 & 0.399 \\
        F1 (deepseek-r1)   & 0.120 & 0.292 & 0.952 & 0.362 \\
        F1 (deepseek-v3)   & 0.031 & 0.297 & 0.951 & 0.253 \\
        \midrule
        Prec (gpt-4o-mini) & 0.235 & 0.214 & \textbf{0.912} & 0.411 \\
        Prec (gpt-4o)      & 0.450 & 0.228 & 0.907 & 0.406 \\
        Prec (o1-mini)     & 0.556 & 0.197 & 0.911 & 0.397 \\
        Prec (o1)          & 0.647 & 0.198 & 0.908 & 0.406 \\
        Prec (llama-70B)   & \textbf{0.833} & 0.169 & 0.907 & 0.438 \\
        Prec (llama-405B)  & 0.667 & \textbf{0.236} & 0.911 & 0.450 \\
        Prec (deepseek-r1) & 0.568 & 0.215 & 0.908 & \textbf{0.454} \\
        Prec (deepseek-v3) & 0.500 & 0.209 & 0.908 & 0.410 \\
        \midrule
        Rec (gpt-4o-mini)  & 0.013 & 0.533 & 0.994 & 0.587 \\
        Rec (gpt-4o)       & 0.058 & \textbf{0.630} & 0.985 & \textbf{0.754} \\
        Rec (o1-mini)      & 0.065 & 0.619 & \textbf{1.000} & 0.646 \\
        Rec (o1)           & \textbf{0.106} & 0.415 & 0.994 & 0.527 \\
        Rec (llama-70B)    & 0.016 & 0.376 & 0.987 & 0.308 \\
        Rec (llama-405B)   & 0.006 & 0.381 & 0.987 & 0.359 \\
        Rec (deepseek-r1)  & 0.067 & 0.455 & \textbf{1.000} & 0.302 \\
        Rec (deepseek-v3)  & 0.016 & 0.515 & 0.998 & 0.183 \\
        \bottomrule
    \end{tabular}
    \label{fig:aspect-weakness}
\end{table}

\clearpage

\subsection{Results using accepted papers}
\label{appendix:result_accepted_paper}

\begin{figure}[H]
    \centering
    \includegraphics[width=\textwidth]{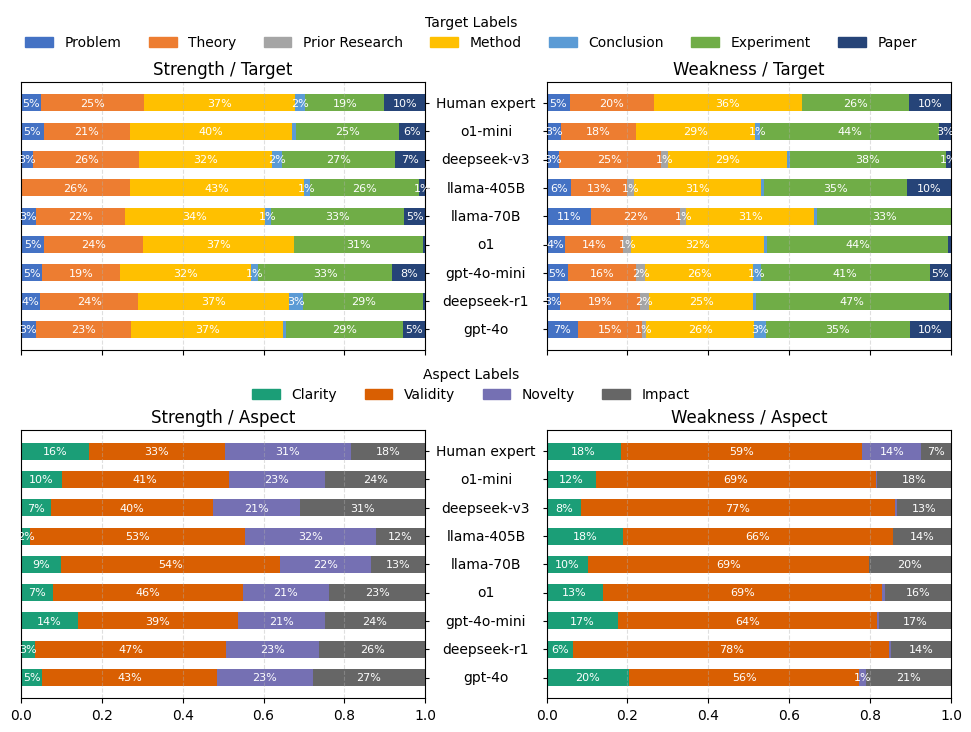}
    \caption{A visualization of focus distributions by target/aspect and strength/weakness for LLMs and human experts using \textit{accepted} papers, in a descending order of KL divergence. We observed a few notable differences in the pattern, compared to the evaluation results using rejected papers. First, there exists a much larger gap in the Weakness-Experiment, meaning that human experts criticize experiments significantly less than LLMs. In strengths, human experts mostly praise Novelty and Impact than Validity, but LLMs tend to praise the Validity the most. We observed the same pattern in Weakness-Novelty, meaning that LLMs neglect the novelty aspect in criticizing the papers.}
\label{fig:accepted_chart}
\end{figure}

\end{onecolumn}

\end{document}